\documentclass[sn-mathphys,Numbered]{sn-jnl}% Math and Physical Sciences Reference Style
\usepackage{graphicx}%
\usepackage{multirow}%
\usepackage{amsmath,amssymb,amsfonts}%
\usepackage{amsthm}%
\usepackage{mathrsfs}%
\usepackage[title]{appendix}%
\usepackage{xcolor}%
\usepackage{textcomp}%
\usepackage{manyfoot}%
\usepackage{booktabs}%
\usepackage{algorithm}%
\usepackage{algorithmicx}%
\usepackage{algpseudocode}%
\usepackage{listings}%
\theoremstyle{thmstyleone}%
\theoremstyle{thmstyletwo}%

\theoremstyle{thmstylethree}%

\usepackage{subcaption}

\begin{document}

\title[Data-driven Nonlinear pROM using Deep Hierarchical VAE \ldots]{Data-driven Nonlinear Parametric Model Order Reduction Framework using Deep Hierarchical Variational Autoencoder}


\author[1]{\fnm{SiHun} \sur{Lee}}\email{leesihun@snu.ac.kr}

\author[1]{\fnm{Sangmin} \sur{Lee}}\email{sj7714@snu.ac.kr}

\author[1]{\fnm{Kijoo} \sur{Jang}}\email{prastins@snu.ac.kr}

\author[2]{\fnm{Haeseong} \sur{Cho}}\email{hcho@jbnu.ac.kr}

\author*[1,3]{\fnm{SangJoon} \sur{Shin}}\email{ssjoon@snu.ac.kr}

\affil[1]{\orgdiv{Department of Aerospace Engineering}, \orgname{Seoul National University}, \orgaddress{\city{Seoul}, \postcode{08226}, \country{Republic of Korea}}}

\affil[2]{\orgdiv{Department of Aerospace Engineering}, \orgname{Jeonbuk National University}, \orgaddress{\city{Jeonju}, \postcode{54896}, \country{Republic of Korea}}}

\affil*[3]{\orgdiv{Institute of Advanced Aerospace Technology}, \orgname{Seoul National University}, \orgaddress{\city{Seoul}, \postcode{08226}, \country{Republic of Korea}}}

\abstract{
A data-driven parametric model order reduction (MOR) method using a deep artificial neural network is proposed. The present network, which is the least-squares hierarchical variational autoencoder (LSH-VAE), is capable of performing nonlinear MOR for the parametric interpolation of a nonlinear dynamic system with a significant number of degrees of freedom. LSH-VAE exploits two major changes to the existing networks: a hierarchical deep structure and a hybrid weighted, probabilistic loss function. The enhancements result in a significantly improved accuracy and stability compared against the conventional nonlinear MOR methods, autoencoder, and variational autoencoder. Upon LSH-VAE, a parametric MOR framework is presented based on the spherically linear interpolation of the latent manifold. The present framework is validated and evaluated on three nonlinear and multiphysics dynamic systems. First, the present framework is evaluated on the fluid-structure interaction benchmark problem to assess its efficiency and accuracy. Then, a highly nonlinear aeroelastic phenomenon, limit cycle oscillation, is analyzed. Finally, the present framework is applied to a three-dimensional fluid flow to demonstrate its capability of efficiently analyzing a significantly large number of degrees of freedom. The performance of LSH-VAE is emphasized by comparing its results against that of the widely used nonlinear MOR methods, convolutional autoencoder, and $\beta$-VAE. The present framework exhibits a significantly enhanced accuracy to the conventional methods while still exhibiting a large speed-up factor.
}




\maketitle

\section{Introduction}\label{section:introduction}

Modern high-fidelity, nonlinear computational analysis is mostly computationally intensive in terms of time and memory. In particular, many multiphysics analysis adopt a partitioned method in which the solvers regarding each type of physics are executed separately. Such an approach also requires computation for the data interpolation among different types of discretization and executes iterative computation within a single time step, demanding even more intensive computation. Consequently, model order reduction (MOR) has been suggested to alleviate the computational time and memory consumption. Two types of MOR frameworks exist: intrusive and non-intrusive. Intrusive MOR depends on the governing equation to construct the reduced bases. Galerkin projection is one of the most widely used approaches which projects an ensemble of the full-order model (FOM) results into the governing equation \cite{rowley2004model, carlberg2011efficient}. However, a parametric analysis may become extremely challenging when the algorithm is not explicitly established as it manipulates the governing equation directly \cite{chen2012blackbox}. Instead, a completely data-driven approach, non-intrusive MOR (NIMOR) may be considered. NIMOR aims to discover the embedded pattern in the FOM dataset and rescale those to a much smaller dimensionality. Unlike intrusive MOR, NIMOR is independent of the governing equation, making it to be extremely versatile. 

Among MOR methods, linear subspace MOR (LS-MOR) has been widely considered as they are mathematically rigorous and efficient. LS-MOR has been successfully employed in fluid dynamics, flow control, structural dynamics, aeroelasticity, and fluid-structure interaction (FSI) \cite{Berkooz1993, ravindran2000reduced, rajagopal2013proper, shane2007structural, kim2018advanced, lee2020time, lee2021parametric, lee2023parametric}. However, LS-MOR may require an excessive number of the subspaces to accurately represent a nonlinear, complex FOM. For example, in complex turbulent fluid flows, proper orthogonal decomposition (POD) extracts its modes with respect to the energy ratio and details are filtered out \cite{carlberg2017galerkin}. Those details are usually excluded because they contain very small energy and the corresponding coefficients are quite random. LS-MOR methods are generally known to be less effective on advection-dominated, sharp-gradient, multiphysics systems, and especially systems with slowly decaying Kolmogorov n-width \cite{kim2022fast, xu2020multi,carlberg2017galerkin,kim2020efficient}.

Recent exponential development in the field of machine learning has enabled neural networks to be used for MOR. Specifically, autoencoder has become a viable nonlinear MOR method where a shallow, well-trained autoencoder with a linear activation function is known to behave similarly to POD \cite{brunton2020machine, kramer1991nonlinear, demers1992non}. Instead of the linear activation functions, many autoencoders adopt nonlinear activation functions, using them to generate nonlinear subspace \cite{kramer1991nonlinear, demers1992non}. Such an autoencoder-based method has been implemented widely to reduce the dimensionality of various engineering problems including fluid dynamics, convection problems, and structural dynamics \cite{gonzalez2018deep, kadeethum2022non, kadeethum2022reduced, kim2022enhanced, xu2020multi, omata2019novel, lee2020model}. However, the performance of an autoencoder as a generative ANN is known to be quite limited \cite{berthelot2018understanding}. The deterministic aspect of its loss function, which was designed to only reconstruct the input, limits autoencoders to generate diverse outputs. Attempts to enhance the generative capability have led to the development of the variational autoencoder (VAE) and generative adversarial network (GAN) \cite{kingma2013auto, goodfellow2014generative}. These methods implement probabilistic loss functions that construct a dense and smooth latent space. Between the two alternatives, VAE is selected for use in this study owing to its stable training property \cite{lee2021parametric}. VAE has been widely studied for use in the field of computer vision but it has also been used to interpolate dynamic systems \cite{lee2021parametric, lee2023parametric}.

VAE in its simplest form, vanilla VAE, is capable of generating data of significantly superior quality compared with the autoencoder. However, VAE commonly suffers from a phenomenon known as posterior collapse, where the generative model learns to ignore a subset of the latent variables \cite{lucas2019understanding}. The posterior collapse was easily alleviated by applying a technique known as Kullback-Leibler divergence (KL divergence) annealing, or $\beta$-VAE  \cite{bowman2015generating, sonderby2016ladder, fu2019cyclical, vahdat2020nvae, burgess2018understanding, higgins2017beta}. Another problem with vanilla VAE is that it is restricted to a shallow network, limiting its expressiveness. Vanilla VAE tends to perform worse as the network becomes deeper due to the loss of long-range correlation and its performance was found to be insufficient when complex data were processed \cite{lee2021parametric, vahdat2020nvae}. Deep hierarchical VAEs, such as the LVAE, IAF-VAE, and NVAE, have been developed to enhance the performance of vanilla VAE \cite{kingma2016improved, sonderby2016ladder, vahdat2020nvae}. These VAEs mainly adopt a type of residual cells that connect the encoder and decoder directly without passing through the latent space. Similar to U-nets, the skip connections allow bidirectional information sharing between the encoder and decoder, thereby preventing the loss of long-range correlation.

Recently, various types of VAEs are being adopted as a nonlinear MOR method owing to their superior generative capability compared to conventional autoencoders. VAEs have been adopted on flow problems \cite{eivazi2020deep, eivazi2022towards, solera2023beta, mrosek2021variational}, transonic flow \cite{kang2022physics, wang2021flow}, numerics \cite{zhu2020information}, biology \cite{kneifl2023low}, brain MRI images \cite{miolane2020learning}, and anomaly detection \cite{wang2019systematic, lee2019process}. While earlier studies adopt the simplest convolutional VAE, many recent studies consider $\beta$-VAE due to its near-orthogonal latent space \cite{burgess2018understanding, higgins2017beta}. Previous studies show that $\beta$-VAE may successfully construct nonlinear subspace, but the majority of networks used in those studies were quite shallow. The use of shallow networks may result in insufficient expressiveness if the input data consists of a large number of DOF and exhibits a complex response.

Instead, a deep hierarchical VAE is proposed, least-squares hierarchical VAE (LSH-VAE) for nonlinear MOR of a dynamic system. LSH-VAE is a very deep hierarchical network that incorporates a modified loss function similar to that of $\beta$-VAE. The deep hierarchical structure enables a very deep, stable network ($>$100 layers) with highly expressive and accurate interpolation results. The modified loss function consists of a hybrid weighted least-squares and Kullback-Leibler divergence function that alleviates posterior collapse and enhances orthogonality of the latent space \cite{kullback1951information, burgess2018understanding, higgins2017beta}. The least-squares error in the loss function is also known to enhance the accuracy when used on the continuous dataset \cite{lee2023parametric}. 

%Particularly, the proposed framework is designed to perform well even with very few initial samples, which is crucial for use in practical engineering environments. 
There has been no report on a very deep VAE ($>$100 layers) implemented for nonlinear MOR. The present framework is validated by solving the following three problems. First, a standard two-dimensional FSI benchmark problem developed by Turek and Hron will be exemplified \cite{turek2006proposal}. Then, the highly nonlinear aeroelastic phenomenon of limit cycle oscillation (LCO) will be considered to examine the accuracy of the proposed framework under nonlinearity. Finally, the flow surrounding a three-dimensional cylinder is to be analyzed to establish the capability of the current framework to accommodate a system with a significantly large number of degrees of freedom. The computational efficiency and accuracy will be assessed as well as comparison to the existing nonlinear MOR methods will be presented.

\section{Machine-Learning Methods}

This section provides the theoretical background of the machine learning methods. Based on the existing convolutional autoencoder and $\beta$-VAE, the formulation of the proposed network, LSH-VAE is presented.

\subsection{Convolutional autoencoder (CAE)}

\begin{figure}[h!]
\centering
\includegraphics[width=0.7\linewidth]{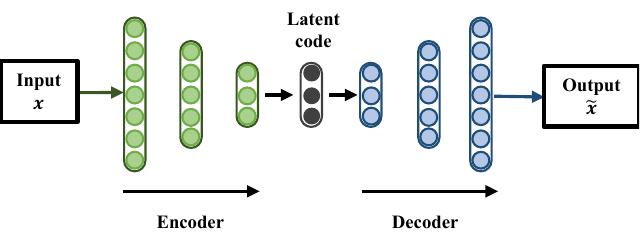}
\caption{Architecture of a typical convolutional autoencoder}
\label{fig:AE_architecture}
\end{figure}

A convolutional autoencoder (CAE) is an ANN that is trained to output data that are similar to its input. The typical architecture of the CAE, shown in Fig. \ref{fig:AE_architecture}, enables the encoder to compress the input data into a smaller latent dimensionality. The decoder then expands the latent code back to its original dimensionality. By training both the encoder and decoder, CAE learns to extract important features of the input dataset. The latent codes contain the embedded features recognized by the CAE that can be used as the reduced bases in the ROM.
The interpolation of data using CAE is conducted by interpolating the latent codes. The interpolated latent code contains the interpolated features, which leads to the interpolation of the input data.

% From here
The loss function of CAE is quite intuitive. CAE takes the input, $x$, and passes it through the encoder, $\Phi$, to obtain the latent vector, $z$. Then, the decoder, $\Psi$, receives the latent vector and generates the output, $y$. The output, $y$, is compared against the input, $x$, using the mean squared error ($MSE$) loss function. In this way, the CAE is trained such that the difference between $y$ and $x$ is reduced, aiming for a more accurate reconstruction of the input. The equations for the encoder and decoder network are presented in Eq. (\ref{eq:AE}), where the loss function is shown in Eq. (\ref{eq:AE_loss}).

\begin{equation} \label{eq:AE}
z=\Phi(x), ~ y = \Psi(z)
\end{equation}

\begin{equation} \label{eq:AE_loss}
L = MSE(\Psi(\Phi(x))-x)
\end{equation}

The simplest form of CAE, known as the vanilla CAE, has been shown to produce unsatisfactory interpolation outcomes \cite{berthelot2018understanding}. Hence, derivatives thereof such as VAE, and GAN may be utilized to enhance the performance.

\subsection{Variational autoencoder (VAE)}

\begin{figure}[h]
\centering
{\includegraphics[width=0.9\textwidth]{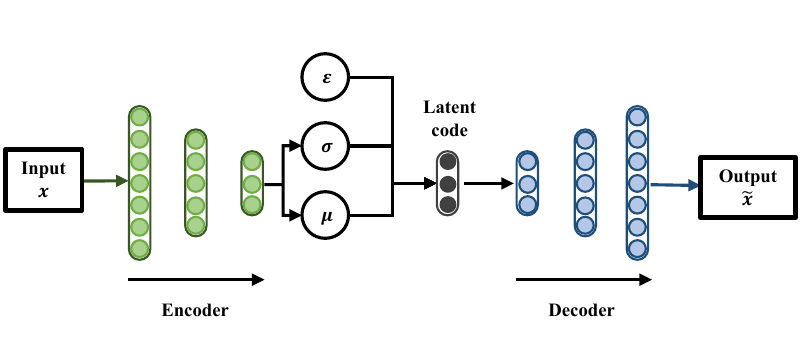}}
\caption{Architecture of a typical VAE\label{fig2}}
\end{figure}

VAE and autoencoder share a similar architecture. The largest difference lies in that the encoder of VAE utilizes probabilistic latent values instead of discrete latent codes. The probabilistic encoder models the latent feature probability distribution. The resultant latent space is continuous and smooth, enabling higher quality generated outcomes. The encoder of VAE extracts the mean, $\mu$, and the variance, $\sigma$, which are used to generate the latent code, $z$. A typical VAE structure can be observed in Figure \ref{fig2}.

VAE aims to efficiently infer the intractable posterior distribution, $p(z \vert x)$. It is performed by adopting an approximate posterior, $q(z \vert x)$, because determining the true posterior is quite challenging. Here, the encoder or inference network is represented by $q(z \vert x)$, whereas the decoder network is denoted as $p(x \vert z)$. 

Kullback-Leibler (KL) divergence is the expectation of the difference between two distributions, which is always a positive value. KL divergence between the approximate and the real posterior is written as Eq. (\ref{eq_1}).

\begin{equation}\label{eq_1}
D_{KL}(q(z  \vert  x)  \vert  \vert  p(z  \vert  x))=-\int q(z  \vert x)\log(\frac{p(z  \vert  x)}{q(z  \vert  x)})dz\ge 0
\end{equation}

Applying Bayes' theorem to Eq. (\ref{eq_1}) yields Eq. (\ref{eq_2}).

\begin{equation}\label{eq_2}
\begin{gathered}
D_{KL}(q(z \vert x) \vert  \vert p(z \vert x)) = -\int q(z \vert x) \log(\frac{p(x \vert z)p(z)}{q(z \vert x)p(x)}) dz \\
= -\int q(z \vert x) \log(\frac{p(x \vert z)p(z)}{q(z \vert x)}) dz + \log p(x)\ge 0
\end{gathered}
\end{equation}

Equation (\ref{eq_2}) can be rewritten as Eq. (\ref{eq_35}). Applying the rules of logarithm to Eq. (\ref{eq_35}) will yield  Eq. (\ref{eq_3}).

\begin{equation}\label{eq_35}
\begin{gathered}
\log p(x) \ge \int q(z \vert x)\log \frac{p(x \vert z)p(z)}{q(z \vert x)}dz
\end{gathered}
\end{equation}

\begin{equation}\label{eq_3}
\begin{gathered}
\log p(x) 
\ge \int q(z \vert x) \log(\frac{p(z)}{q(z \vert x)})dz + \int q(z \vert x)\log p(x \vert z) dz\\
\ge \mathbb{E}_{q(z  \vert  x)}[\log p(x \vert z)]-D_{KL}(q(z  \vert  x) \vert \vert p(z)) 
\end{gathered}
\end{equation}

The right hand side of Eq. (\ref{eq_3}) is the evidence lower bound (ELBO). VAE aims to maximize ELBO which maximizes the logarithmic probability of the data by proxy. Following the convention of minimizing the loss function, the right hand side of Eq. (\ref{eq_3}) is converted as Eq. (\ref{eq9}), which is the goal of VAE.

\begin{equation}\label{eq9}
\min[ -\mathbb{E}_{q(z  \vert  x)}[\log p(x  \vert  z)]+ D_{KL}(q(z  \vert  x) \vert \vert p(z)) ]
\end{equation}

The goal of VAE is to minimize both the reconstruction and KL divergence loss. In Eq. (\ref{eq9}), the first term corresponds to the reconstruction loss and the second term corresponds to KL divergence loss. KL divergence loss enforces the decoder (approximate posterior) to become similar to the inverse of the encoder.

The loss function in Eq. (\ref{eq9}) has to be differentiable to minimize it during the training. Usually, KLD term can be integrated analytically \cite{kingma2013auto}; however, the reconstruction loss is not directly differentiable. To enforce the reconstruction loss to be differentiable, the reparameterization technique is adopted \cite{kingma2013auto}. 

First, Gaussian sampled random noise, $\varepsilon$ will be introduced. The latent code $z$, is formulated as shown in Eq. (\ref{eq_z}), introducing the mean and standard deviation to the equation.

\begin{equation}\label{eq_z}
z=\mu+(\sigma\times\varepsilon),~ \varepsilon\sim N(0,1)
\end{equation}

Since the latent code is formulated as Eq. (\ref{eq_z}), KL divergence in Eq. (\ref{eq9}) is rewritten as Eq. (\ref{eq91}), assuming the posterior and prior follow the Gaussian distribution.

\begin{equation}\label{eq91}
D_{KL}(q(z\vert x)\vert \vert p(z))  = \frac{1}{2}\sum(\sigma^2+\mu^2-(\log(\sigma^2)+1))
\end{equation}

 The latent code with the reparameterization technique enforces the latent space to be stochastically determined. The reparameterization enables the reconstruction loss to be differentiable by Monte Carlo method. For further details and step-by-step derivation of the VAE loss function, reference can be found in works by Kingma and Odaibo \cite{kingma2013auto, odaibo2019tutorial}.

\subsection{Least-squares hierarchical variational autoencoder (LSH-VAE)}

\begin{figure}[h]
\centering
{\includegraphics[width=0.6\textwidth]{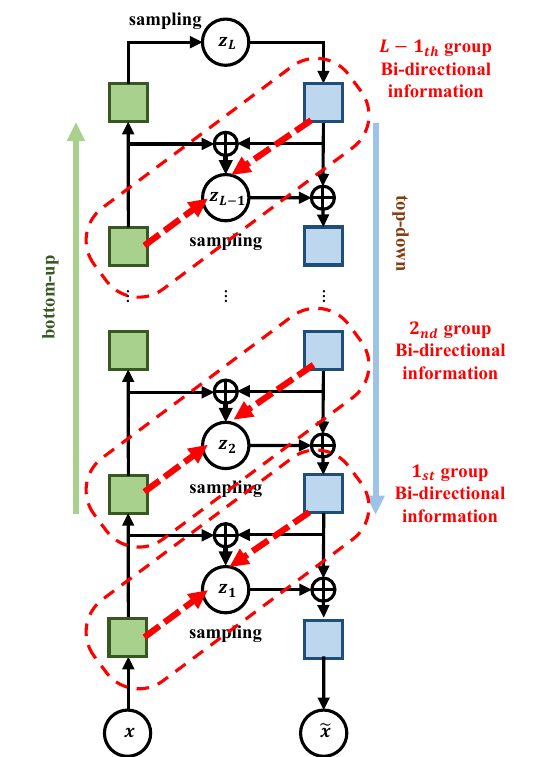}}
\caption{Hierarchical structure of LSH-VAE\label{fig_ladder}}
\end{figure}

Conventional vanilla VAE is limited to shallow networks due to vanishing gradients and the loss of long-range correlation. However, shallow networks may lack expressiveness on complex systems with a significant number of DOFs. In this study, a deep VAE with a hierarchical structure is proposed to enhance the performance. Specifically, to alleviate the loss of long-range correlation and stabilize the training process of a very deep network. The hierarchical structure creates direct passages between the earlier layers of the encoder and the latter layers of the decoder, circumventing the middle layers. Those direct passages enable bidirectional information sharing between the encoder and decoder network. The bidirectional information enables the earlier layers of the VAE to greatly affect the outcome, thus, alleviating the loss of long-range correlation. The diagram in Fig. \ref{fig_ladder} shows the hierarchical structure of LSH-VAE.

In the hierarchical VAE, the latent variables are divided into $L$ groups. By the divided latent dimension, the prior and posterior distributions are rewritten as in Eq. (\ref{eq10}) and Eq. (\ref{eq11}).

\begin{equation}\label{eq10}
\begin{gathered}
p(z)=p\left(z_{L}\right) \prod_{i=1}^{L-1} p\left(z_{i}  \vert  z_{i+1}\right)\\
q(z  \vert  x)=q\left(z_{1}  \vert  x\right) \prod_{i=2}^{L} q\left(z_{i}  \vert  z_{i-1}\right)
\end{gathered}
\end{equation}

\begin{equation}\label{eq11}
\begin{gathered}
p(z_{i}  \vert  z_{i+1})=\mathcal{N}(z_{i}  \vert  \mu(z_{i+1}), \sigma^{2}(z_{i+1}))\\ \quad p(z_{L})=\mathcal{N}(z_{L}  \vert  0, I)\\
q(z_{i}  \vert  z_{i-1})=\mathcal{N}(z_{i}  \vert  \mu(z_{i-1}), \sigma^{2}(z_{i-1}))\\ \quad q(z_{1}  \vert  x)=\mathcal{N}(z_{1}  \vert  \mu(x), \sigma^{2}(x))
\end{gathered}
\end{equation}

The loss function for hierarchical VAE is shown in Eq. (\ref{eq12}), which is obtained by computing the KL divergence separately for each group. By breaking down the KL divergence into groups, bidirectional information flows are created between the inference and generative network. Detailed descriptions about the deep hierarchical structure of VAE can be found in \cite{sonderby2016ladder}.

\begin{equation}\label{eq12}
\begin{gathered}
\min [ -\mathbb{E}_{q(z  \vert  x)}[\log p(x  \vert  z)]+ D_{K L}(q(z  \vert  x) \ \vert  p(z))\\
+\sum_{i=1}^{L-1} \mathbb{E}_{q(z_{<i}  \vert  x)}[D_{K L}(q(z_{i}  \vert  z_{<i}, x) \ \vert  p(z_{i}  \vert  z_{>i}))]]
\end{gathered}
\end{equation}

The present LSH-VAE adopts hierarchical structures motivated by LVAE, IAF-VAE, and NVAE \cite{kingma2016improved, sonderby2016ladder, vahdat2020nvae}. The latent codes in the hierarchical VAE are formed by both bottom-up and top-down information. The latent codes of each of the groups output shared information (from the encoder and decoder) to the next decoder block. Because the information of the encoder and decoder network is shared via latent code, the network delivers higher performance.

Upon the hierarchical structure, LSH-VAE implements a hybrid weighted loss function. The loss function consists of the mean squared error (MSE) and KL divergence instead of conventional binary cross entropy. The use of MSE as a reconstruction error has been known to be successful for continuous datasets \cite{lee2023parametric}. The loss function of LSH-VAE is shown in Eq. (\ref{eq:LSH-VAE}), where the coefficients $\alpha$ and $\beta$ denote the weights of the MSE and KL divergence, respectively.

\begin{equation}\label{eq:LSH-VAE}
\begin{gathered}
\min _{\phi, \theta} [\alpha MSE(x, \tilde{x})+ \beta D_{K L}(q(z  \vert  x) \ \vert  p(z))\\
+\sum_{i=1}^{L-1}\mathbb{E}_{q(z_{<i}  \vert  x)}[\beta D_{K L}(q(z_{i}  \vert  z_{<i}, x) \ \vert  p(z_{i}  \vert  z_{>i}))]]
\end{gathered}
\end{equation}

Usually, the weights $\alpha$ and $\beta$ are set to be $\alpha / \beta_{target}\approx 10^6$. During the training, $\alpha$ is a fixed value whereas $\beta$ is a variable that varies with respect to the epochs. The variable $\beta$ is implemented to prevent posterior collapse in which some latent variables become inactive. This method is known as KL-annealing or $\beta$-VAE, where $\beta$ is formulated as Eq. (\ref{KLD}) \cite{bowman2015generating}.

\begin{equation}\label{KLD}
\beta = 
\begin{cases}
1\times 10^{-4}\beta_{target} &\text{if epoch $<0.3n_{epochs}$}\\
\beta_{target}\frac{epoch}{n_{epochs}}&\text{if epoch $>0.3n_{epochs}$}
\end{cases}
\end{equation}

During the training, $\beta$ is assigned a low value at the start such that LSH-VAE behaves as an autoencoder. During the first few epochs, input data will be mapped on the latent space. Beyond a few prescribed epochs, $\beta$ will be gradually ramped up such that LSH-VAE may behave as a VAE, generating smooth latent space.

\section{Present Framework} 

\subsection{Architecture of the least-squares hierarchical VAE (LSH-VAE)}

\begin{figure}[h!]
\centering
{\includegraphics[width=0.8\textwidth]{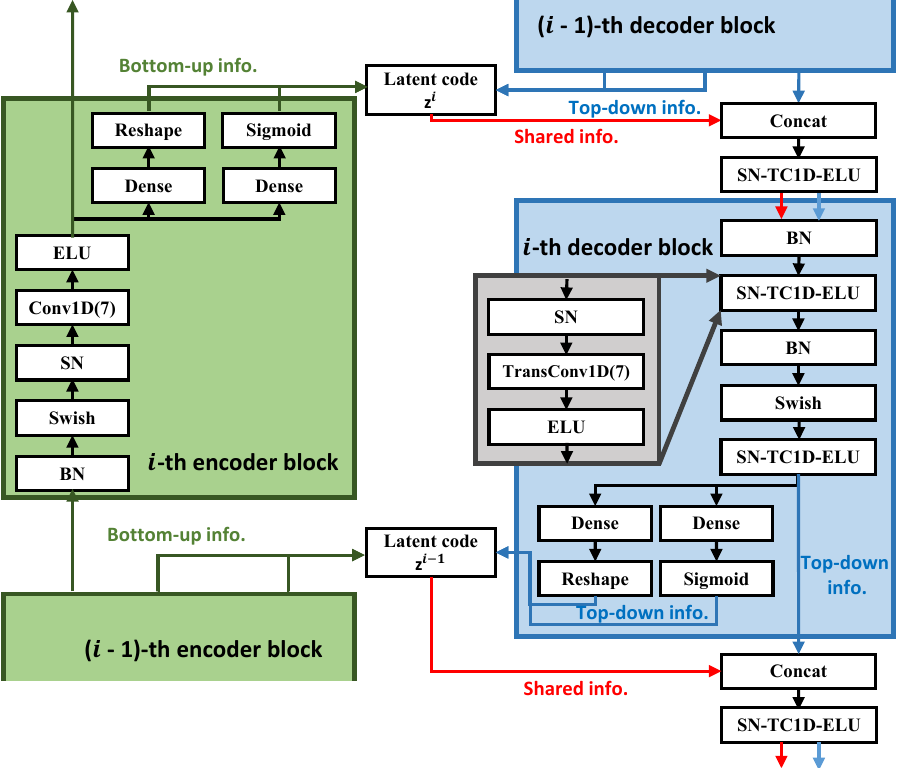}}
\caption{Detailed architecture of the encoder and decoder blocks of LSH-VAE\label{fig:LSH-VAE_block}}
\end{figure}

LSH-VAE adopts a one-dimensional (1D) convolutional layer to accommodate the transient response of the unstructured grids. The use of a 1D convolutional layer enables the temporal continuity of the physical variables to be considered. 
The encoder and decoder of the LSH-VAE consist of the blocks discussed in the previous section, where a detailed schematic of these blocks is shown in Fig. \ref{fig:LSH-VAE_block}.

Being a deep neural network (DNN), LSH-VAE encoder and decoder blocks are composed of stacks of multiple layers. These layers consist of the following layers: spectral normalization (SN), 1D convolution, dense, exponential linear unit (ELU), Swish, and batch normalization (BN). Swish, and ELU nonlinear activation functions are chosen as their continuous derivatives enhance the stability of a DNN \cite{clevert2015fast}. The LSH-VAE implements a normalization-activation sequence instead of the conventional activation-normalization sequence. Such sequence is known to deliver benign performance empirically when used before the convolutional computation \cite{vahdat2020nvae}. The output of the encoder block is branched in three ways. The first branch connects to the input of the next block and the remaining two branches form $\mu$, and $\sigma$. The encoder latent code is formulated by reparameterizing $\mu$, and $\sigma$.
The reparameterized latent code and ELU layer infer bottom-up information transfer, shown in green in Fig. \ref{fig:LSH-VAE_block}. 

In the current configuration, the decoder network is significantly deeper and more complex than the encoder network. The deep decoder network enables an expressive output when accompanied by a system with many DOFs. The decoder network receives two inputs: top-down information from the predecessor decoder block and encoder-decoder shared information from the latent code. Through a series of layers, the decoder outputs top-down information, shown in blue. The decoder block generates the decoder latent code and input for the next block. The encoder latent code and the decoder latent code are added to generate shared latent code, $z^i$. The shared latent code contains both top-down and bottom-up information, enabling bidirectional information sharing.

\subsection{Preprocessing dataset}

Acquiring many FOM samples may be quite cumbersome. In particular, many-queried FOM computations are extremely time-consuming if FOM is highly nonlinear, includes multiphysics, and involves a significant number of DOFs. Acquiring those FOM data through experiments and simulations is considered prohibitive for computational, financial reasons. Instead, data augmentation is considered to sample sparsely and expand the amount of training data. A larger amount of training data improves the generalization of ANN and thus enhances the accuracy. Similar to the data augmentation typically performed on images, the pre-acquired FOM results are processed using the following three methods. First, temporal data are resampled by shortening the timestep, i.e. frequency elongation. Then, the training data are augmented by changing the amplitude and adding a random number within the bound of $\pm$30\% for every epoch. Training the ANN using the augmented data ensures that the ANN is effectively trained against a very large dataset, resulting in a high-performance network.

\subsection{LSH-VAE training and interpolation}

The current framework performs MOR directly on FOM results. The LSH-VAE employs 1D convolutional layers which requires a three-dimensional input of the format $\left(batch, sequence, channel\right)$. % In the format, the $sequence$ dimension is the convolutional dimension, which takes the neighboring data into consideration. The $channel$ dimension, however, is independent of its neighboring data.
In the current configuration, the temporal continuity of the FOM results is considered in the convolutional dimension. The resultant input composition of LSH-VAE becomes $\left(batch, N_t, N_{DOF}\right)$, where $N_t$ denotes the number of time steps and $N_{DOF}$ denotes the number of DOFs in the dynamic system. LSH-VAE receives such input and compresses it into latent vectors via the encoder. The dimensionality change throughout LSH-VAE is expressed in Eq. \ref{eq15}, where $N_i$ represents the latent dimension in the $i$-th latent group. The total latent dimension, $\sum N_i$ is much smaller than the FOM dimension, achieving MOR.

\begin{equation}\label{eq15}
\left(, N_t, N_{DOF}\right)\xrightarrow{\text{encoder}}
\left(, \sum N_{i}\right)
\xrightarrow{\text{decoder}}\left(, N_t, N_{DOF}\right)
\end{equation}

\begin{algorithm}
\caption{Training of LSH-VAE}\label{alg1}
\begin{algorithmic}
\Require {Hyperparameters, $\alpha_{LR}, \beta_{1}, \beta_{2}, N_A$, $q^{enc}_{\phi}, p^{dec}_{\theta}$}
\State {$x=N(v^i) =\bar{v}^{i}\rightarrow[-0.7,0.7]\left(i=1, \cdots, N_{DOF}\right)$}
\State {$x_{train} = [x,~$R$(x)_1,~$R$(x)_2,~...~,~$R$(x)_{N_A}]$ (resampling aug.)}
\For {$i=1, N_{epoch}$}
\State {$x_{train} = 0.3\alpha\times (x_{train}+0.3\beta), ~ \alpha,\beta\sim\mathcal{N}\left(0, 1\right)$ (amplitude, offset aug.)}
\For {$j=1, L$}
\If {$j=1$}
\State {$q^{enc}_{\phi,1}\left(x_{train}\right)=\{u_{1}, \mu_{q,1}, \sigma_{q,1}\}$}
\State {$z_1=\mu_{q,1}+\sigma_{q,1}\varepsilon, \varepsilon\sim\mathcal{N}\left(0, 1\right)$}
\ElsIf {$1<j<L$}
\State {$q^{enc}_{\phi,j}\left(u_{j-1}\right)=\{u_{j}, \mu_{q,j}, \sigma_{q,j}\}$}
\State {$z_j=\mu_{q,j}+\sigma_{q,j}\varepsilon, \varepsilon\sim\mathcal{N}\left(0, 1\right)$}
\ElsIf {$j=L$}
\State {$q^{enc}_{\phi,L}\left(u_{L-1}\right)=\{\mu_{q,L}, \sigma_{q,L}\}$}
\EndIf
\EndFor
\State {$z_{L}=\mu_{q,L}+\sigma_{q,L}\times\varepsilon, ~~\varepsilon\sim\mathcal{N}\left(0, 1\right)$}
\For {$k=L,-1,1$}
\If {$k=L$}
\State {$p^{dec}_{\theta,k}\left(z_{L}\right)=\{d_{k},\mu_{p,k}, \sigma_{p,k}\}$}
\State {$z_{L-1}^{dec}=\mu_{p,L}+\sigma_{p,L}\varepsilon, \varepsilon\sim\mathcal{N}\left(0, 1\right)$}
\State {$z_{L-1} = z_{L-1}+z_{L-1}^{dec}$}
\ElsIf {{$1\leq k<L$}}
\State {$p^{dec}_{\theta,k}\left(d_{k+1}, z_{k}\right)=\{d_{k},\mu_{p,k}, \sigma_{p,k}\}$}
\State {$z_{k-1}^{dec}=\mu_{p,k}+\sigma_{p,k}\varepsilon, \varepsilon\sim\mathcal{N}\left(0, 1\right)$}
\State {$z_{k-1} = z_{k-1}+z_{k-1}^{dec}$}
\EndIf
\EndFor
\State{$\mathcal{L}_{LSH-VAE}\rightarrow$Eq. (7), where $x = x_{train}, \tilde{x} = d_1$}
\State Adamax{$\left(\nabla_{\phi, \theta} \mathcal{L}_{LSH-VAE}, \alpha_{LR}, \beta_{1}, \beta_{2}\right) \rightarrow \phi, \theta$}
\EndFor
\end{algorithmic}
\end{algorithm}

The training algorithm for LSH-VAE is shown in Algorithm \ref{alg1}. The algorithm starts by normalizing the physical variables of interest, $v$. $v$ is normalized to the range of [-0.7, 0.7] for each DOF by the normalizing function, $N()$. The normalized variable is then augmented by resampling for $N_A$ instances. Then, the training dataset, $x_{train}$ is constructed by concatenating the original normalized variable with the augmented ones. The training dataset of the network becomes, $x_{train} = [x,~$R$(x)_1,~$R$(x)_2,~...~,~$R$(x)_{N_A}]$, where R$(x)_n$ denotes the resampled normalized variable of interest.

The training dataset is further augmented for amplitude and offset. The amplitude and offset augmentation is performed by using random values for every epoch. The network receives a different input in every epoch, enabling the network to be trained against a very large dataset. After the data augmentation is completed, the encoder and the decoder networks are trained. After the decoder is trained, the loss function can be obtained by Eq. \ref{eq:LSH-VAE}. The training of LSH-VAE is optimized by the Adamax optimizer, which has shown good performance compared with the conventional Adam and SGD optimizers. 

Generative ANNs usually require latent vectors to be sought. This is required owing to the probabilistic formulation that is used to parameterize the latent vector. However, we empirically found that sufficient epochs and a small number of parameters obviate the need for latent searching. In this study, rather than attempting latent searching, the latent vectors are calculated by the mean value from the encoder network directly. 

Upon acquiring the latent vectors, $slerp$ interpolation is performed to collect the targeted latent vector. The latent space created by VAEs is in the form of a well-structured, multi-dimensional hypersphere, which enables complex operation by vector arithmetic \cite{larsen2016autoencoding}. It is possible since the reparameterization trick introduces Gaussian random number, which attributes to the vector length and angle in the latent hypersphere. The $slerp$ interpolation shown in Algorithm \ref{alg3} not only interpolates the rotation angle of vectors, but it also interpolates the arc length. Such $slerp$ interpolation enables the latent vectors to be interpolated following the path of the complex latent manifold. The use of $slerp$ interpolation has been widely accepted for performing latent interpolation \cite{white2016sampling, agustsson2017optimal}.

\begin{algorithm}
\caption{Interpolation of latent codes}\label{alg3}
\begin{algorithmic}
\Require {$N()$ from Algorithm \ref{alg1}, $z^{i}$}
\State{Set the target parametric value, $p_{tgt}$, and two adjacent values, $p_{1}$ and $p_{2}$}
\State{Find the ratio, $k$, such that $p_{tgt}=k\times p_{1}+\left(1-k\right)\times p_{2}$}
\State{$z_{p_{tgt}}=slerp(z_{p_{1}}, z_{p_{2}}, k)
= cos^{-1}(p_1\cdot p_2)= p_1\frac{\sin(1-k)\theta}{\sin\theta}+ p_2\frac{\sin k \theta}{\sin\theta}$}
\State{$v_{p_{tgt}}=N^{-1}(f^{dec}_{\phi}\left(z_{p_{tgt}}\right))$}
\end{algorithmic}
\end{algorithm}

\section{Numerical results} 

This section presents the numerical results obtained by the proposed framework. First, the framework is applied to solve a FSI benchmark problem previously developed by Turek and Hron \cite{turek2006proposal}. The accuracy of the current method is evaluated and compared against that obtained by the conventional nonlinear MOR, CAE. Then, the proposed framework is examined on a wing section that undergoes limit cycle oscillation (LCO). LCO analysis is performed to evaluate the accuracy of the proposed framework on the nonlinear multiphysics phenomenon. Last, the applicability of LSH-VAE to a system with many DOFs is demonstrated by analyzing a three-dimensional fluid flow.

The numerical results presented in this paper are obtained by intentionally sampling a small number of initial FOM results. Sparse sampling is performed because ANN replicating its training data often leads to enough accuracy when the sampling is performed densely. In addition, sparse sampling is attempted as dense and iterative computations on a nonlinear system with many DOFs are rather unrealistic. 

For all of the results, the same LSH-VAE network is used for each variable of interest. The hyperparameters used for the training are shown in Table. \ref{hyper_TH}. In Table \ref{hyper_TH}, the first value for the latent dimension criterion denotes the latent dimension in which the interpolation is performed. The latter value denotes the latent dimension used for information sharing between the encoder and decoder network. LSH-VAE used for the following numerical results consists of 7 encoder and decoder blocks, with a total of 107 layers. While detailed optimization of the hyperparameters would yield better accuracy, such procedure is not performed to emphasize the generability of the framework. However, different batch sizes are used considering the number of DOF, limited by the VRAM of GPU.

\begin{center}
\begin{table}[h]
\centering
\caption{Hyperparameters of LSH-VAE\label{hyper_TH}}
\begin{tabular}{cccc}
\toprule
\textbf{Criterion} & \textbf{Value}&\textbf{Criterion} & \textbf{Value}\\
\midrule
Epochs & 5,000&Latent dim.&[32, 8]\\
$\alpha$ & $1\times10^6$&Number of filters & [64,32,16,8,4,2,1]\\
$\beta_{target}$ & 1&$N_A$&5\\
Learning rate & $1\times10^{-3}$&&\\
\bottomrule
\end{tabular}
\end{table}
\end{center}

For all of the results presented in this paper, computations are carried out on AMD 3950X CPU to obtain the FOM results. ANN are trained using NVIDIA GeForce GTX 3090 GPU.

\subsection{Turek-Hron FSI benchmark}

\subsubsection{Description of the analysis}

\begin{figure}[h]
\centering
\includegraphics[width=0.6\linewidth]{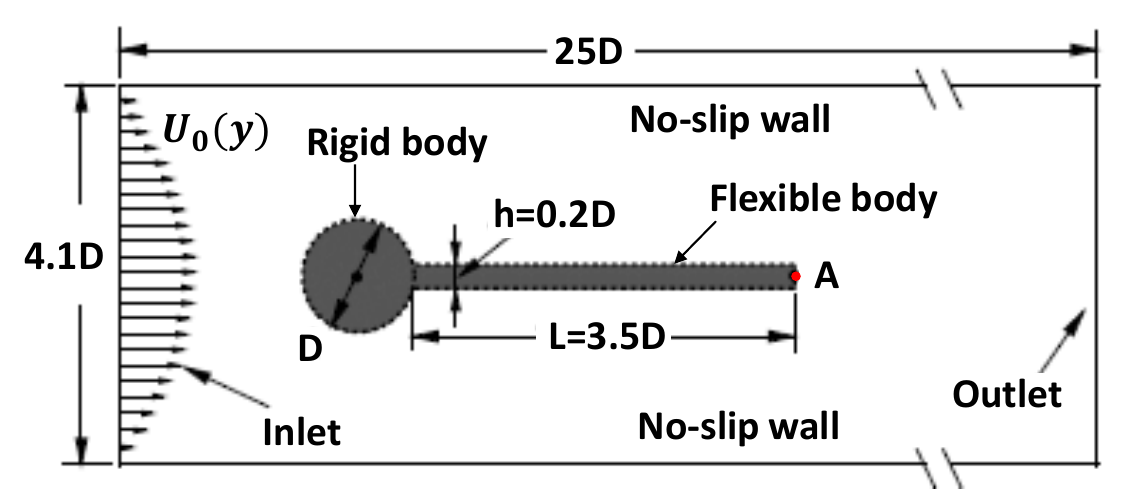}
\caption{Analysis setup of the Turek-Hron FSI problem}
\label{fig:TH_setup}
\end{figure}

The widely accepted FSI benchmark developed by Turek and Hron is described in this section \cite{turek2006proposal}. The benchmark problem consists of a rigid cylinder with a diameter of 0.1 m and a highly flexible tail. The fluid flowed from the inlet to the outlet with laminar separation occurring behind the cylinder. Von K\`{a}rm\`{a}n vortex street created by the flow separation excites the tail, which exhibits a large deflection. A hyperbolic inlet profile is used to consider the no-slip initial wall boundary condition at the upper and lower computational domain. A detailed schematic regarding the analysis is shown in Fig. \ref{fig:TH_setup}.%The original benchmark problem was solved at an inflow velocity of 1m/s, which showed a fully converged displacement of $\pm80.1$mm at point $A$. 

The current framework requires a few parametric initial FOM samples to extract the embedded patterns. For Turek-Hron FSI benchmark problem, seven initial FOM results are collected. The inflow speed was selected as a parameter and speeds ranging from 0.7 $m/s$ to 1.3 $m/s$, in 0.1 $m/s$ intervals were sampled. The FOM samples are analyzed using Navier-Stokes computational fluid dynamics (CFD) and finite element method (FEM) two-way FSI analysis provided in the commercial software, ANSYS. The flow field is discretized by 29,788 CFD nodes and the flexible body is discretized by 954 FEM nodes.

The ensemble of FOM results is constructed by collecting 2 s of the fully converged response in intervals of 0.01 s. The pre-acquired FOM ensemble is then subjected to interpolation by LSH-VAE shown in Table \ref{hyper_TH}. After the training of LSH-VAE is completed, the latent code is interpolated. In the present case, the target parameter is selected as the unseen inflow speed of 0.95$m/s$. The latent code corresponding to 0.95$m/s$ is acquired by the slerp interpolation shown in Algorithm \ref{alg3}. The interpolated latent code is then decoded by the decoder network where the resultant interpolated variables are generated.

\subsubsection{Accuracy and efficiency}

\begin{figure*}[h!]
\captionsetup[sub]{font=normal}
\centering

\begin{subfigure}[b]{0.48\textwidth}
%\centering
\includegraphics[width = \textwidth]{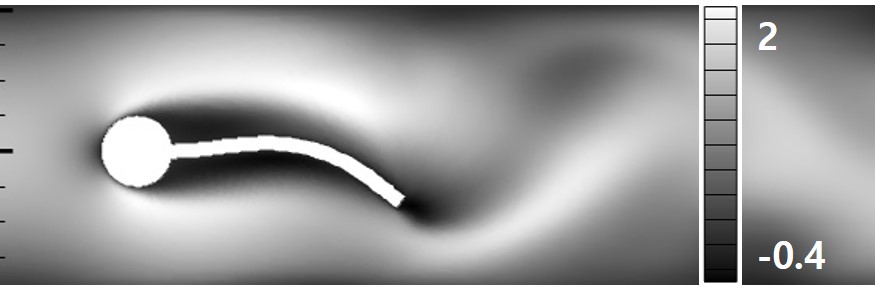}
\caption{LSH-VAE, $u$}
\end{subfigure}
\begin{subfigure}[b]{0.48\textwidth}
\centering
\includegraphics[width = \textwidth]{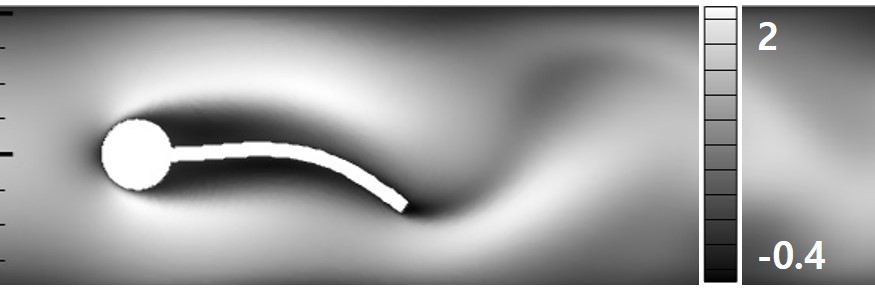}
\caption{FOM, $u$}
\end{subfigure}

\begin{subfigure}[b]{0.48\textwidth}
\centering
\includegraphics[width = \textwidth]{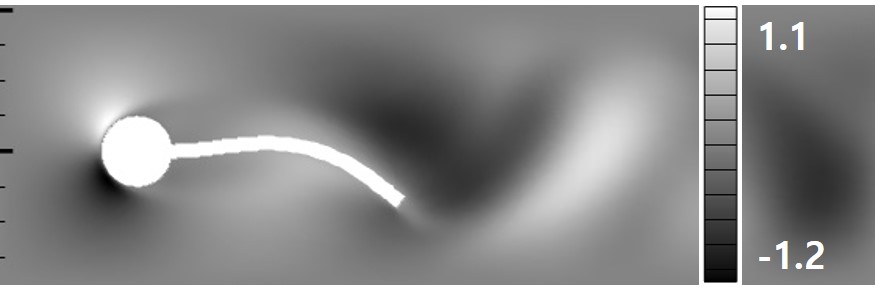}
\caption{LSH-VAE, $v$}
\end{subfigure}
\begin{subfigure}[b]{0.48\textwidth}
\centering
\includegraphics[width = \textwidth]{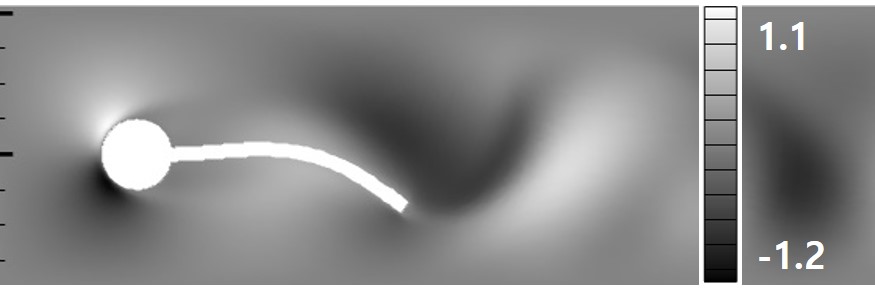}
\caption{FOM, $v$}
\end{subfigure}

\begin{subfigure}[b]{0.48\textwidth}
\centering
\includegraphics[width = \textwidth]{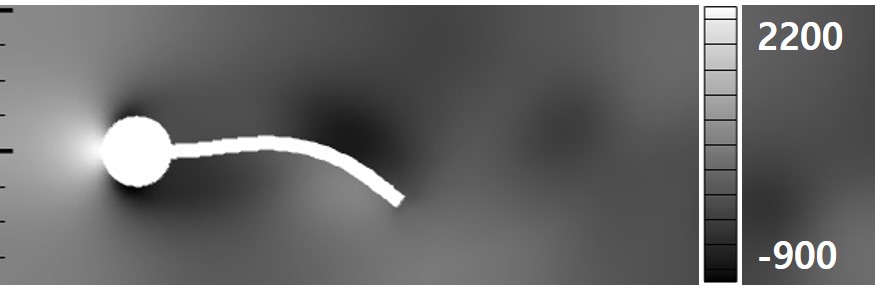}
\caption{LSH-VAE, $p$}
\end{subfigure}
\begin{subfigure}[b]{0.48\textwidth}
\centering
\includegraphics[width = \textwidth]{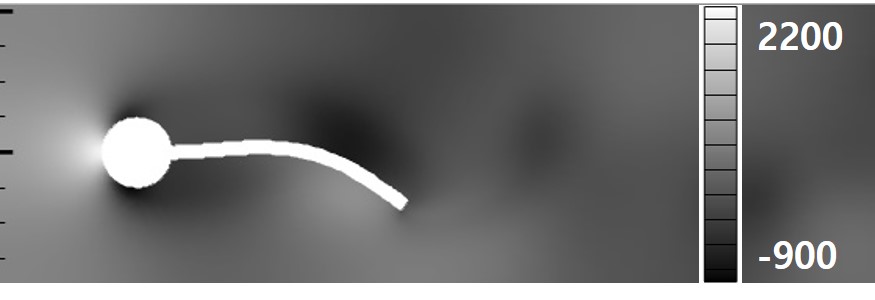}
\caption{FOM, $p$}
\end{subfigure}

\caption{Original and interpolated FSI field for Turek-Hron FSI problem at $t=2$ s}\label{fig_TH_field}
\end{figure*}

The accuracy of the current framework is assessed by comparing the results of the ROM against those obtained with the FOM. Five physical variables, $dX, dY, u, v$, and $p$ are considered for interpolation in this case. Among them, the first two variables denote the grid deformation in x- and y-direction. Using the interpolated variables, the interpolated FSI field will be constructed. The interpolated FSI field and FOM are shown in Fig. \ref{fig_TH_field}.

\begin{figure*}[h!]
\captionsetup[sub]{font=normal}
\centering
\begin{subfigure}[b]{0.32\textwidth}
\centering
\includegraphics[width = \textwidth]{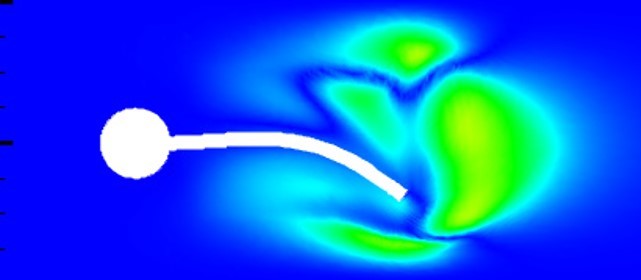}
\caption{LSH-VAE, $dX$}
\end{subfigure}
\begin{subfigure}[b]{0.32\textwidth}
\centering
\includegraphics[width = \textwidth]{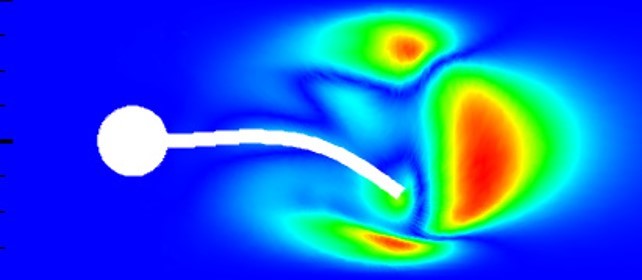}
\caption{CAE, $dX$}
\end{subfigure}
\begin{subfigure}[b]{0.32\textwidth}
\centering
\includegraphics[width = \textwidth]{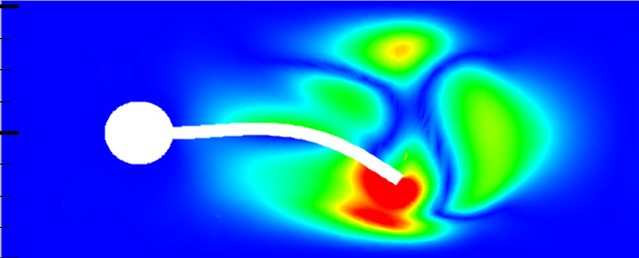}
\caption{$\beta-$VAE, $dX$}
\end{subfigure}

\begin{subfigure}[b]{0.32\textwidth}
\centering
\includegraphics[width = \textwidth]{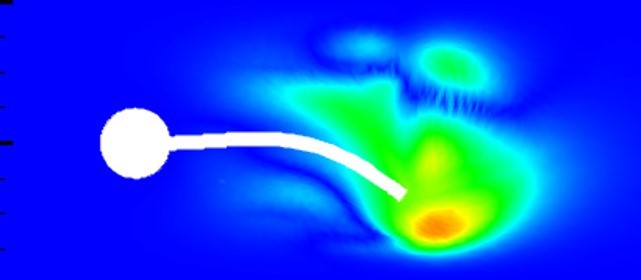}
\caption{LSH-VAE, $dY$}
\end{subfigure}
\begin{subfigure}[b]{0.32\textwidth}
\centering
\includegraphics[width = \textwidth]{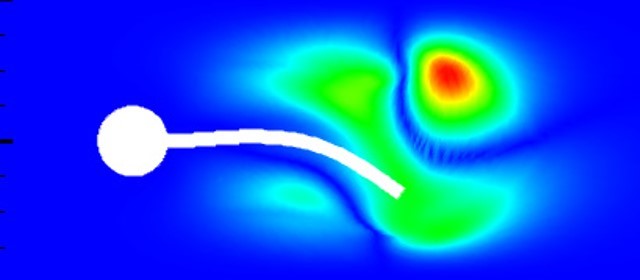}
\caption{CAE, $dY$}
\end{subfigure}
\begin{subfigure}[b]{0.32\textwidth}
\centering
\includegraphics[width = \textwidth]{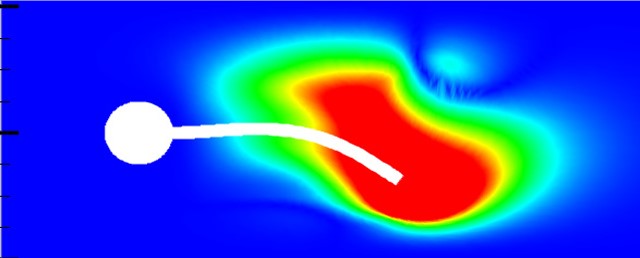}
\caption{$\beta-$VAE, $dY$}
\end{subfigure}

\begin{subfigure}[b]{0.32\textwidth}
\centering
\includegraphics[width = \textwidth]{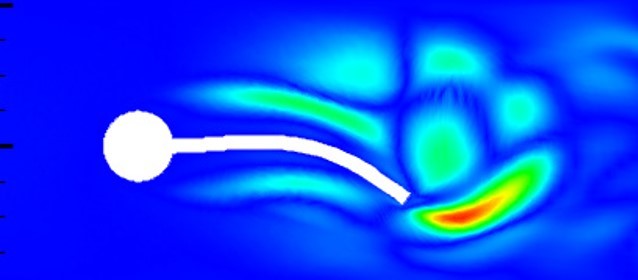}
\caption{LSH-VAE, $u$}
\end{subfigure}
\begin{subfigure}[b]{0.32\textwidth}
\centering
\includegraphics[width = \textwidth]{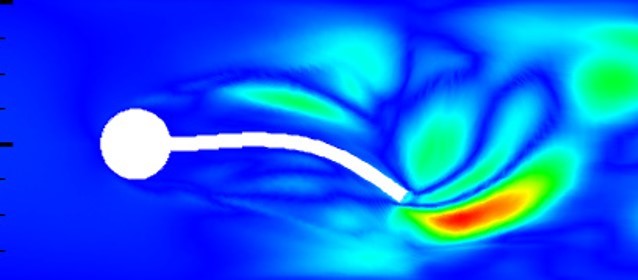}
\caption{CAE, $u$}
\end{subfigure}
\begin{subfigure}[b]{0.32\textwidth}
\centering
\includegraphics[width = \textwidth]{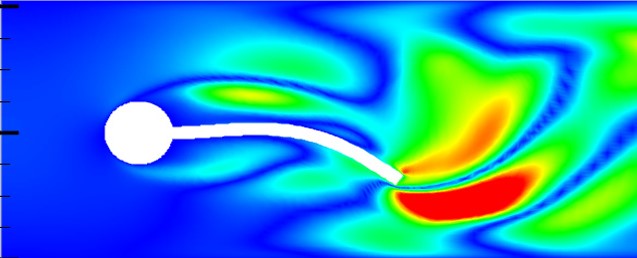}
\caption{$\beta-$VAE, $u$}
\end{subfigure}

\begin{subfigure}[b]{0.32\textwidth}
\centering
\includegraphics[width = \textwidth]{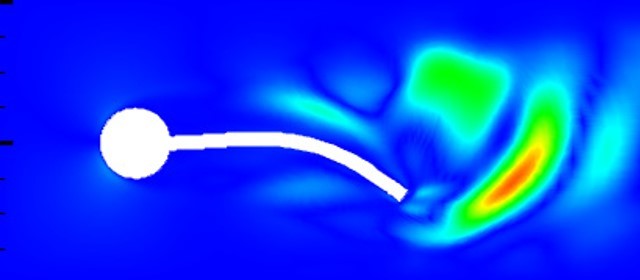}
\caption{LSH-VAE, $v$}
\end{subfigure}
\begin{subfigure}[b]{0.32\textwidth}
\centering
\includegraphics[width = \textwidth]{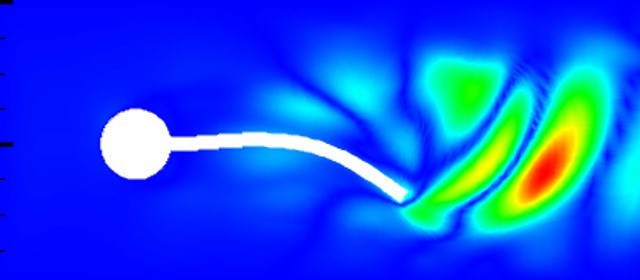}
\caption{CAE, $v$}
\end{subfigure}
\begin{subfigure}[b]{0.32\textwidth}
\centering
\includegraphics[width = \textwidth]{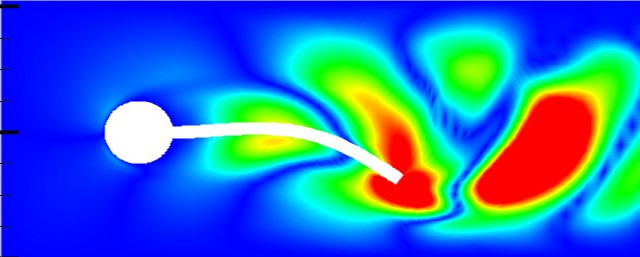}
\caption{$\beta-$VAE, $v$}
\end{subfigure}

\begin{subfigure}[b]{0.32\textwidth}
\centering
\includegraphics[width = \textwidth]{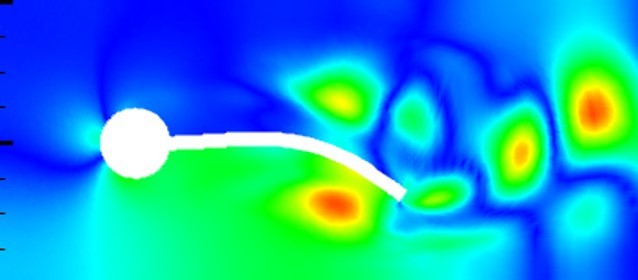}
\caption{LSH-VAE, $p$}
\end{subfigure}
\begin{subfigure}[b]{0.32\textwidth}
\centering
\includegraphics[width = \textwidth]{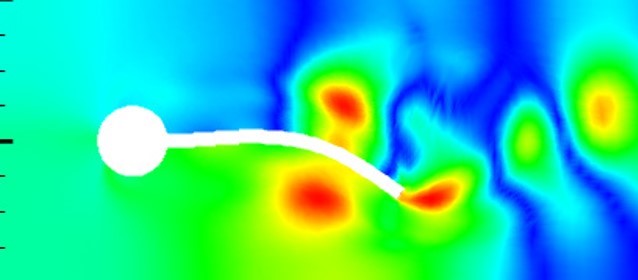}
\caption{CAE, $p$}
\end{subfigure}
\begin{subfigure}[b]{0.32\textwidth}
\centering
\includegraphics[width = \textwidth]{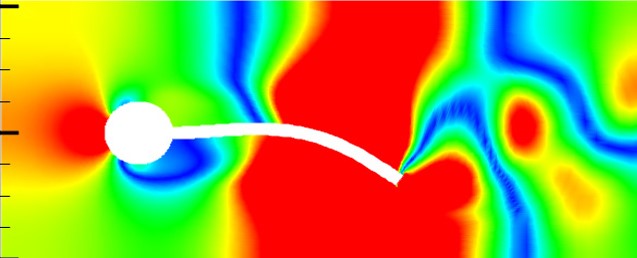}
\caption{$\beta-$VAE, $p$}
\end{subfigure}

\caption{Discrepancy contour of the interpolated FSI field using LSH-VAE, CAE, and $\beta-$VAE at $t=2$ s}\label{fig_TH_diff}
\end{figure*}

Evaluation of the results shown in Fig. \ref{fig_TH_field} verifies that the proposed framework is reasonably accurate. Subsequently, the accuracy of LSH-VAE is compared against that of CAE and $\beta$-VAE. For comparison, the CAE and $\beta$-VAE networks are constructed using the same hyperparameters that were used for LSH-VAE. The comparison between CAE, $\beta$-VAE, and LSH-VAE is performed by comparing the extent to which their results differed from those of FOM. The discrepancy contours of various networks are shown in Fig. \ref{fig_TH_diff}. The minimum and maximum of each variable are matched for the respective variable.

Overall, LSH-VAE exhibits the smallest discrepancy while $\beta$-VAE performs the worst. Interestingly, the regions that exhibit a relatively larger discrepancy are found to be quite similar for all of the networks. This is caused by the finite number of latent dimensions considered in the generative networks. Small details of FOM would have been neglected in the finite latent representation, which lead to the discrepancy in the similar areas. Another one to note is that the pressure contour of CAE and $\beta$-VAE shows a considerably larger discrepancy compared against that by LSH-VAE. This is caused by the large variation between the maximum and minimum values of the pressure. The inability of CAE and $\beta$-VAE to generate an expressive output is considered to be the reason for small details being neglected by large variations.

\begin{figure}[h!]
\centerline{\includegraphics[scale=0.5]{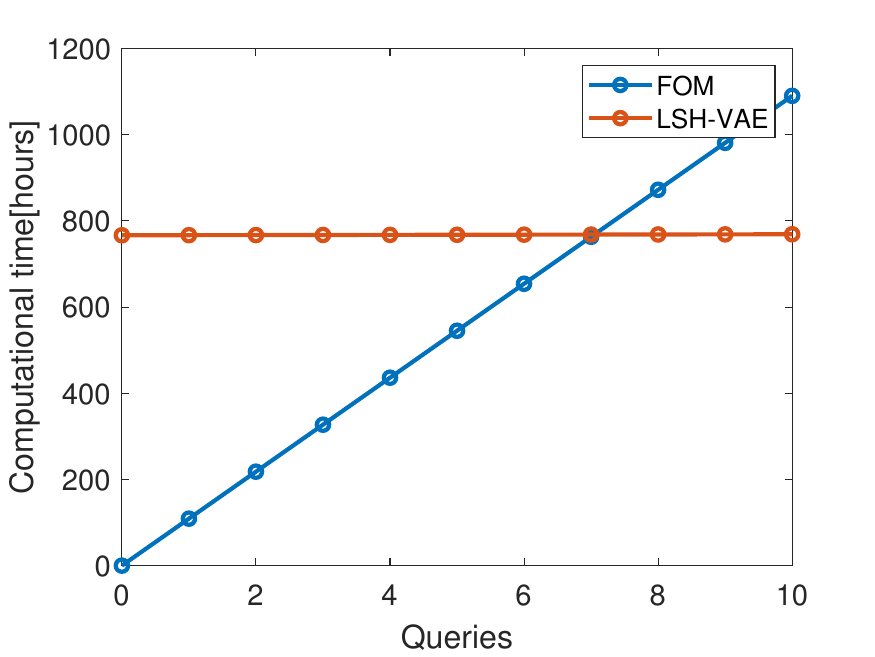}}
\caption{Computational time in terms of the parametric queries for Turek-Hron FSI problem\label{fig15}}
\end{figure}

Then, the efficiency of the proposed framework is assessed. The computational procedures for the proposed framework comprise four stages and the computational time required for each stage is listed in Table \ref{tab5}. For Turek-Hron FSI problem, each FOM query requires 109.0 h whereas the online stage consumes 0.11 h. The proposed framework therefore exhibits a speed-up factor of 990 for each unseen parametric estimation. The expected computational time in terms of the number of computations is shown in Fig. \ref{fig15}.

\begin{center}
\begin{table}[h!]
\centering
\caption{Computational time requirements for Turek-Hron FSI problem\label{tab5}}
\begin{tabular}{llc}
\toprule
&Procedure &Computational time[h]\\ 
\midrule
\multirow{3}{*}{Offline} & FOM & 763.1\\ 
& {Algorithm 1} & 3.51\\
\cmidrule{2-3} 
& Total offline stage & 766.6\\ 
\midrule
\multirow{3}{*}{Online} & Algorithm 2& $<$0.01\\ 
& Variable reconstruction/write & 0.11\\ 
\cmidrule{2-3} 
& Total online stage & 0.11\\ \midrule
&Total sum & 766.7\\ 
%\multicolumn{2}{l}{Total sum} & 82.7\\ 
\bottomrule
\end{tabular}
\end{table}
\end{center}

\subsection{Limit cycle oscillations}

\subsubsection{Description of the analysis}

Limit cycle oscillation (LCO) is a nonlinear periodic oscillation with limited amplitude on an aerodynamic surface. LCO of an aircraft is a highly nonlinear FSI phenomenon that is caused by nonlinearities in both the fluid and structure. Typical causes of LCO include flow separation, transonic shock, geometric nonlinearity, and nonlinear stiffness of the control surface. For an aircraft, LCO may result in structural fatigue in the wings, thus requiring high-fidelity analysis for safety. 
%To consider the nonlinearity in the fluid dynamics, typical LCO analysis implements Euler or Navier-Stokes equation. 
%The amplitude and frequency of LCO usually increase with respect to the flight speed. 
During the design stage of an aircraft, iterative LCO analysis is performed to satisfy the vibration criterion. Such parametric LCO analysis is considered to be quite cumbersome and tedious as it is highly nonlinear and involves many DOFs. In this section, the proposed framework is used to conduct a simplified nonlinear parametric LCO analysis of a wing section.

\begin{figure}[h!]
\centering
\includegraphics[width=0.6\linewidth]{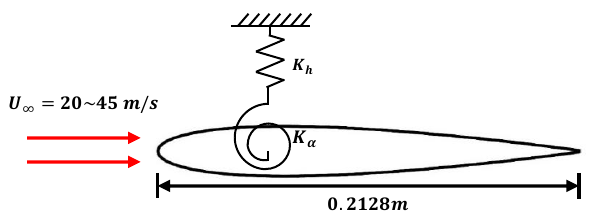}
\caption{Analysis setup of the LCO}
\label{LCO_setup}
\end{figure}

The wing section considered in this analysis is derived from that reported by O'Neil et al. \cite{o1998aeroelastic}. In it, a two-dimensional wing section was constrained by the pitch and heave springs as shown in Fig. \ref{LCO_setup}. The pitch and heave stiffnesses are nonlinear in their cubic terms, which are expressed in Eq. \ref{eqLCO}. LCO is caused by the cubic stiffness in the structure and LCO is observed at the inflow stream speed of 15.5 m/s to 50 m/s.

\begin{equation}\label{eqLCO}
\begin{gathered}
K_{\alpha} = 2.57(\alpha+500\alpha^3)\\
K_{h} = 0.09(h+2860h^3))
\end{gathered}
\end{equation}

The inflow speed is chosen as the parameter in this analysis. The initial FOM samples are collected by adjusting the inflow speed from 20 $m/s$ to 45 $m/s$ in increments of 5 $m/s$. The relevant flow field is discretized by 19,381 nodes and solved using the commercial Navier-Stokes solver, ANSYS. The initial FOM samples are obtained by collecting 2 s of the fully converged response in intervals of 0.01 s. The FOM ensemble is subjected to MOR and interpolation by LSH-VAE. 

After LSH-VAE is trained, the latent code for the desired parameter is acquired via slerp interpolation. The target parameter is an unseen inflow speed of 32.5 $m/s$, and the corresponding latent code is interpolated using Algorithm \ref{alg3}. The interpolated latent code is then decoded by the decoder and the interpolated FSI field is generated.

\subsubsection{Evaluation of accuracy and efficiency}

\begin{figure*}[h!]
\captionsetup[sub]{font=normal}
\centering

\begin{subfigure}[b]{0.48\textwidth}
%\centering
\includegraphics[width = \textwidth]{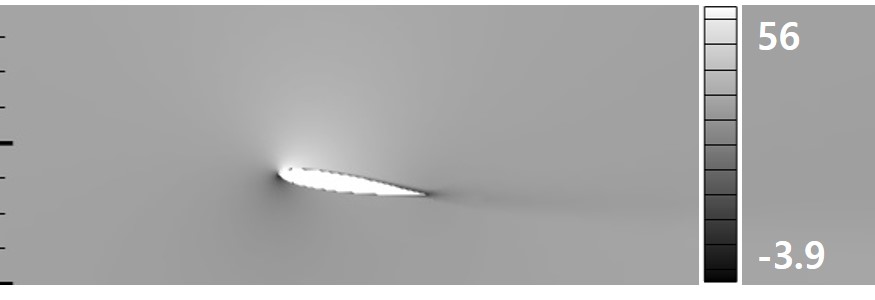}
\caption{LSH-VAE, $u$}
\end{subfigure}
\begin{subfigure}[b]{0.48\textwidth}
\centering
\includegraphics[width = \textwidth]{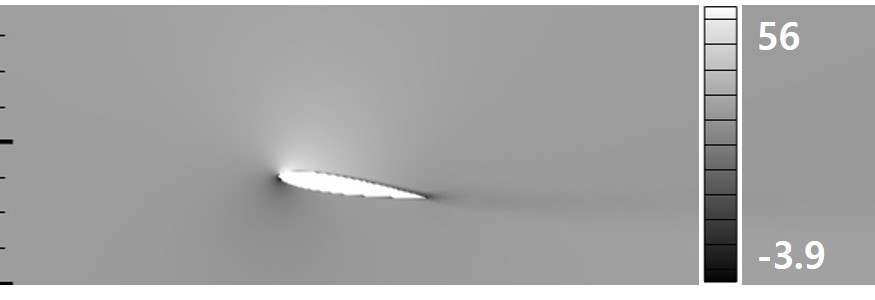}
\caption{FOM, $u$}
\end{subfigure}

\begin{subfigure}[b]{0.48\textwidth}
\centering
\includegraphics[width = \textwidth]{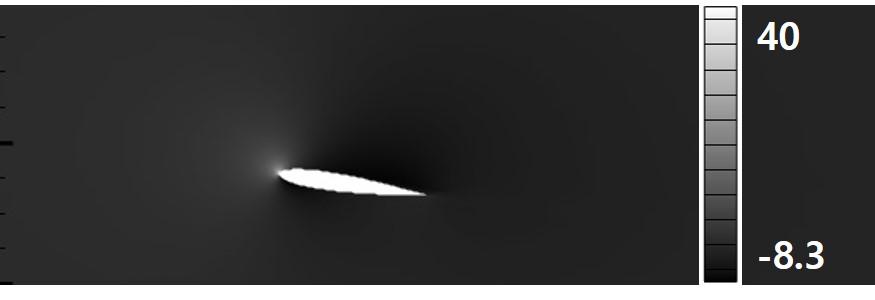}
\caption{LSH-VAE, $v$}
\end{subfigure}
\begin{subfigure}[b]{0.48\textwidth}
\centering
\includegraphics[width = \textwidth]{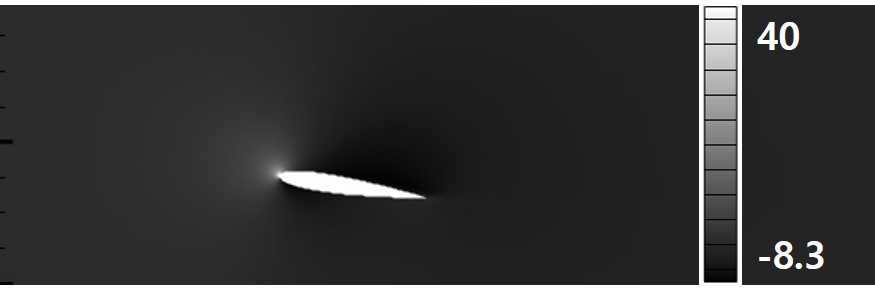}
\caption{FOM, $v$}
\end{subfigure}

\begin{subfigure}[b]{0.48\textwidth}
\centering
\includegraphics[width = \textwidth]{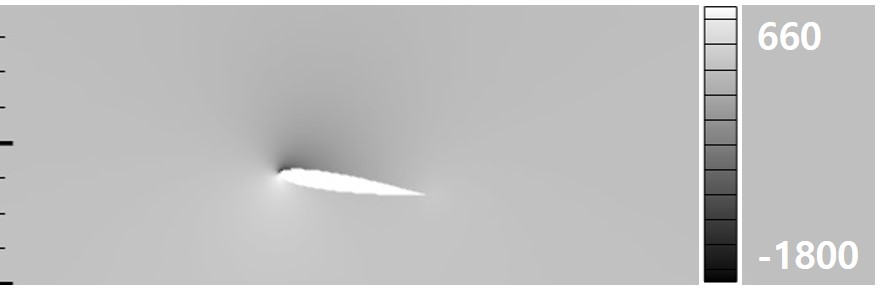}
\caption{LSH-VAE, $p$}
\end{subfigure}
\begin{subfigure}[b]{0.48\textwidth}
\centering
\includegraphics[width = \textwidth]{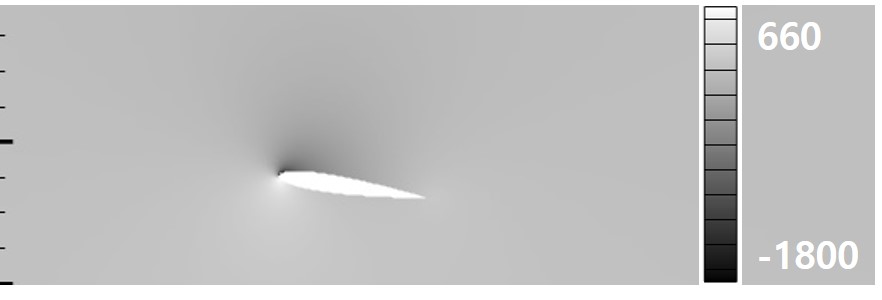}
\caption{FOM, $p$}
\end{subfigure}

\caption{Original and interpolated FSI field for LCO at $t=2$ s}\label{fig_LCO_field}
\end{figure*}

The accuracy of LSH-VAE is assessed by comparing the ROM results against those produced by FOM. In this case, the five physical variables discussed in the previous section were considered. The interpolated variables were used to generate the FSI field, where the interpolated FSI field and FOM are shown in Fig. \ref{fig_LCO_field}.

\begin{figure*}[h!]
\captionsetup[sub]{font=normal}
\centering
\begin{subfigure}[b]{0.32\textwidth}
\centering
\includegraphics[width = \textwidth]{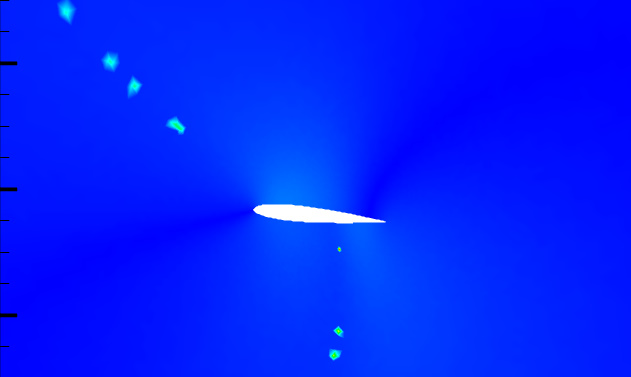}
\caption{LSH-VAE, $dX$}
\end{subfigure}
\begin{subfigure}[b]{0.32\textwidth}
\centering
\includegraphics[width = \textwidth]{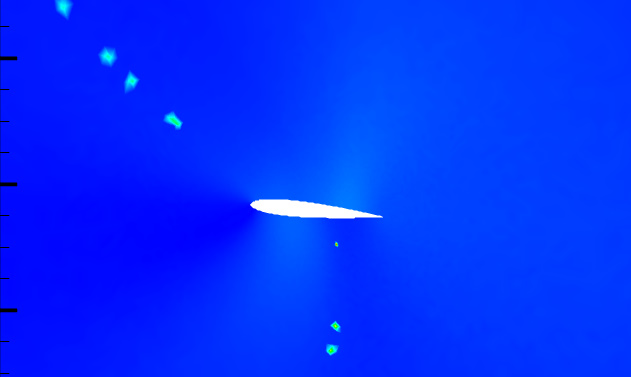}
\caption{CAE, $dX$}
\end{subfigure}
\begin{subfigure}[b]{0.32\textwidth}
\centering
\includegraphics[width = \textwidth]{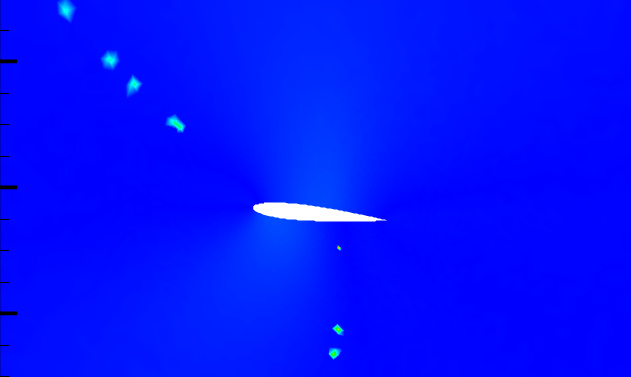}
\caption{$\beta-$VAE, $dX$}
\end{subfigure}

\begin{subfigure}[b]{0.32\textwidth}
\centering
\includegraphics[width = \textwidth]{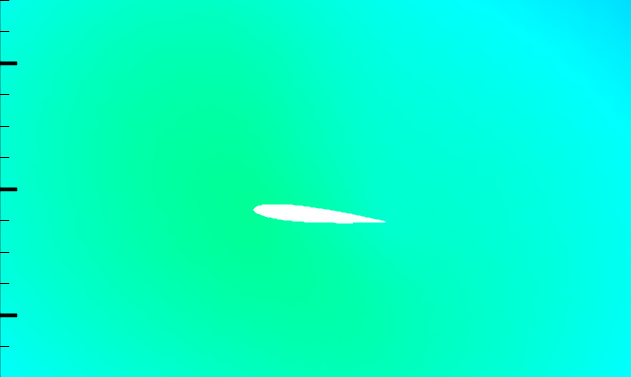}
\caption{LSH-VAE, $dY$}
\end{subfigure}
\begin{subfigure}[b]{0.32\textwidth}
\centering
\includegraphics[width = \textwidth]{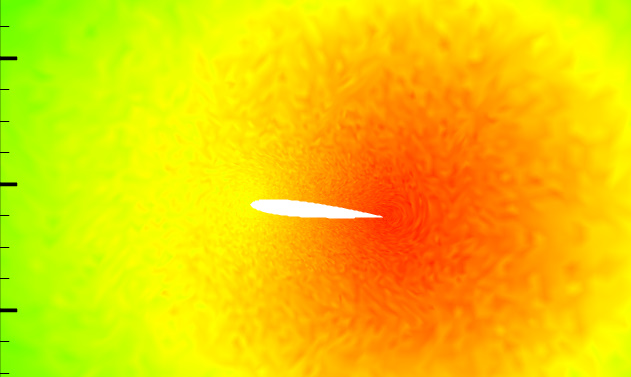}
\caption{CAE, $dY$}
\end{subfigure}
\begin{subfigure}[b]{0.32\textwidth}
\centering
\includegraphics[width = \textwidth]{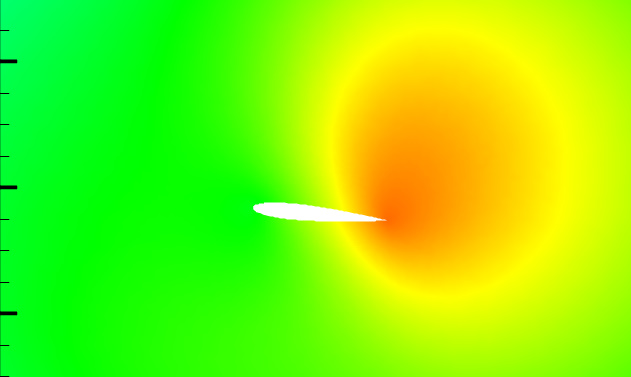}
\caption{$\beta-$VAE, $dY$}
\end{subfigure}

\begin{subfigure}[b]{0.32\textwidth}
\centering
\includegraphics[width = \textwidth]{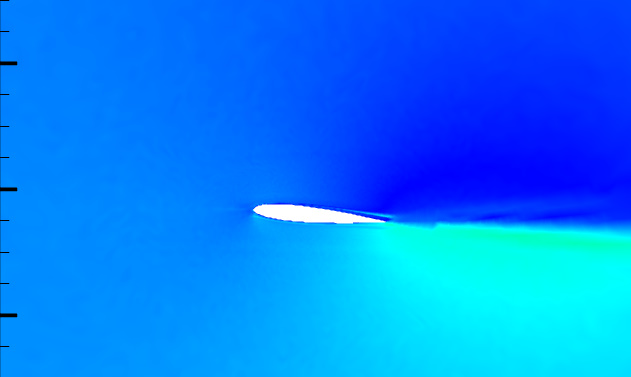}
\caption{LSH-VAE, $u$}
\end{subfigure}
\begin{subfigure}[b]{0.32\textwidth}
\centering
\includegraphics[width = \textwidth]{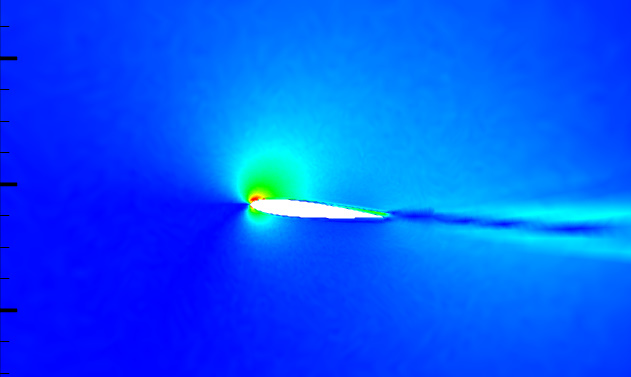}
\caption{CAE, $u$}
\end{subfigure}
\begin{subfigure}[b]{0.32\textwidth}
\centering
\includegraphics[width = \textwidth]{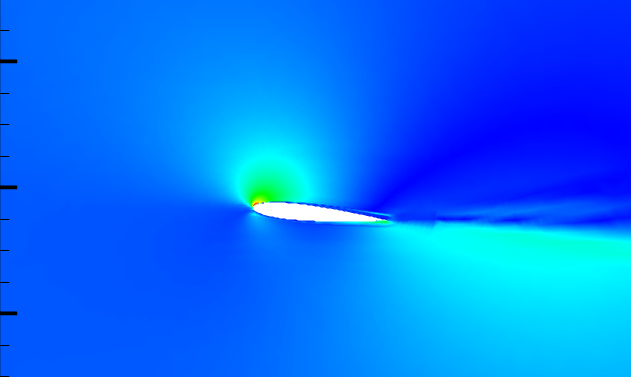}
\caption{$\beta-$VAE, $u$}
\end{subfigure}

\begin{subfigure}[b]{0.32\textwidth}
\centering
\includegraphics[width = \textwidth]{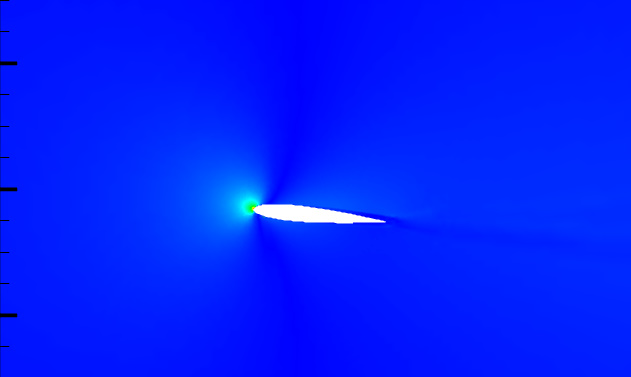}
\caption{LSH-VAE, $v$}
\end{subfigure}
\begin{subfigure}[b]{0.32\textwidth}
\centering
\includegraphics[width = \textwidth]{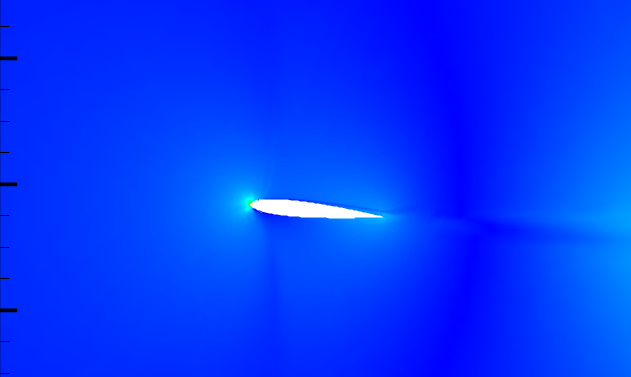}
\caption{CAE, $v$}
\end{subfigure}
\begin{subfigure}[b]{0.32\textwidth}
\centering
\includegraphics[width = \textwidth]{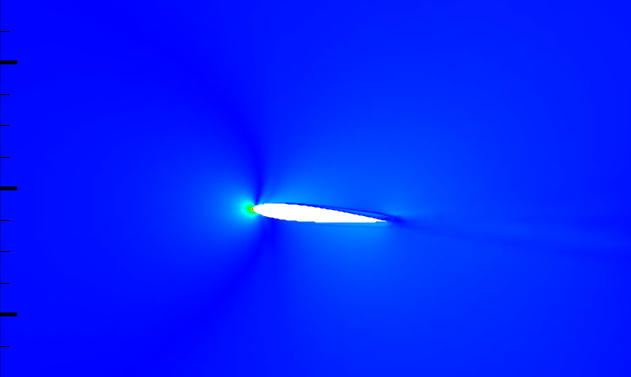}
\caption{$\beta-$VAE, $v$}
\end{subfigure}

\begin{subfigure}[b]{0.32\textwidth}
\centering
\includegraphics[width = \textwidth]{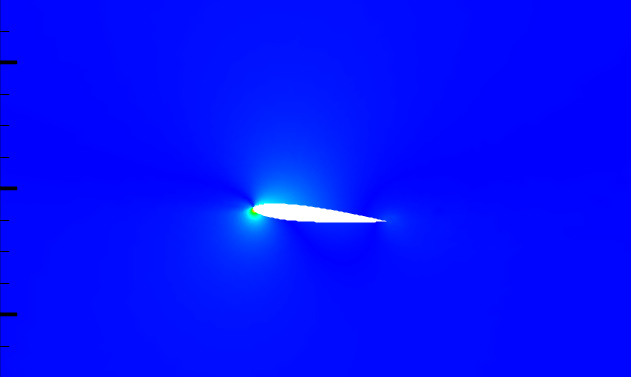}
\caption{LSH-VAE, $p$}
\end{subfigure}
\begin{subfigure}[b]{0.32\textwidth}
\centering
\includegraphics[width = \textwidth]{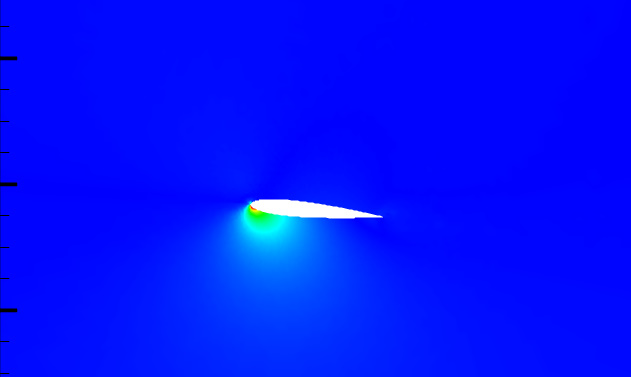}
\caption{CAE, $p$}
\end{subfigure}
\begin{subfigure}[b]{0.32\textwidth}
\centering
\includegraphics[width = \textwidth]{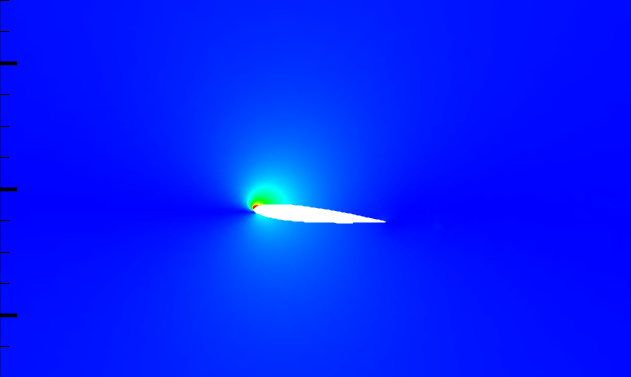}
\caption{$\beta-$VAE, $p$}
\end{subfigure}

\caption{Discrepancy contour of the interpolated FSI field using LSH-VAE, CAE, and $\beta-$VAE at $t=2$ s}\label{fig_LCO_diff}
\end{figure*}

In Fig. \ref{fig_LCO_field}, the interpolated FSI field constructed by LSH-VAE is found to be accurate. Then, the accuracy of LSH-VAE is compared against that of CAE and $\beta$-VAE. The discrepancy contours between LSH-VAE, CAE, and $\beta$-VAE are shown in Fig. \ref{fig_LCO_diff}. The minimum and maximum of the variable are each matched for the same variable.

Similar to Turek-Hron problem, LSH-VAE exhibits the smallest discrepancy. However in this case, $\beta$-VAE performed better than CAE. For $dX$, all networks exhibit a similar discrepancy, as the wing section is constrained in x-direction. Only the pitching motion affects the deformation of surrounding grids in x-direction, resulting in a small variation. $dY$, however, shows different behavior. The discrepancy is spread evenly as the wing heaves and LSH-VAE shows a significantly reduced discrepancy. Another important point to note is that the discrepancy regarding the pressure is quite small. This is due to the stagnation point which creates a concentrated high-pressure region.

The efficiency of the proposed framework is also assessed. The computational time required for each stage is summarized in Table \ref{tab:LCO}. The offline FOM computation required 280.1 h including six initial FOM sample computations. LSH-VAE training required 3.52 h for the five variables of interest, resulting in a total offline stage of 283.6 h. For the online stage, FSI field reconstruction and saving to disk requires the most time as it requires 0.06 h. The present framework exhibits a speed-up factor of 660 for each unseen parametric estimation. 
The expected computational time in terms of the unseen parametric queries is shown in Fig. \ref{fig:LCO_query}.

\begin{center}
\begin{table}[h!]
\centering
\caption{Computational time requirements for LCO\label{tab:LCO}}
\begin{tabular}{llc}
\toprule
&Procedure &Computational time[h]\\ 
\midrule
\multirow{3}{*}{Offline} & FOM & 280.1\\ 
& {Algorithm 1} & 3.52\\
\cmidrule{2-3} 
& Total offline stage & 283.6\\ 
\midrule
\multirow{3}{*}{Online} & Algorithm 2& $<$0.01\\ 
& Variable reconstruction/write & 0.06\\ 
\cmidrule{2-3} 
& Total online stage & 0.06\\ \midrule
&Total sum & 283.7\\ 
%\multicolumn{2}{l}{Total sum} & 82.7\\ 
\bottomrule
\end{tabular}
\end{table}
\end{center}

\begin{figure}[h!]
\centerline{\includegraphics[scale=0.5]{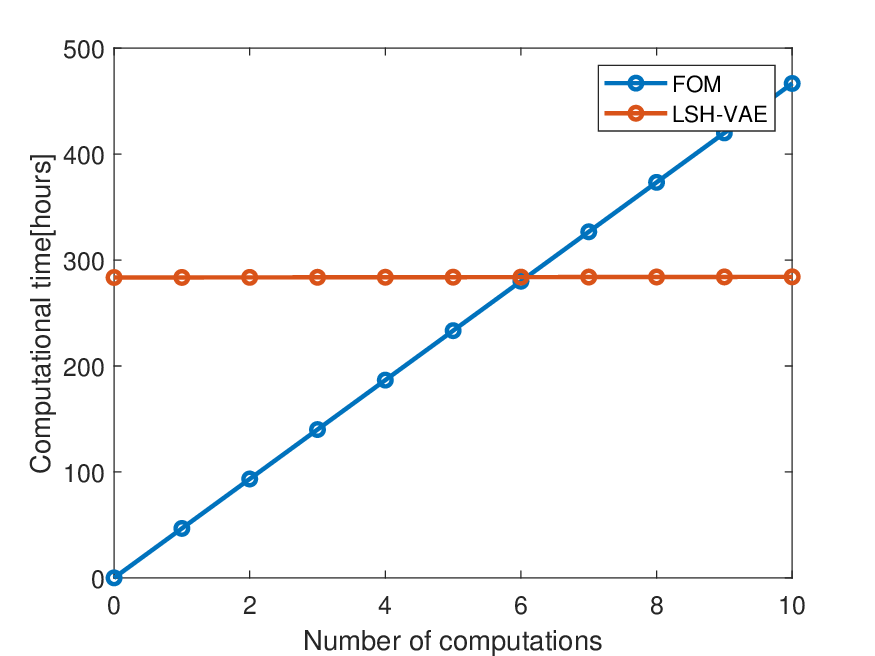}}
\caption{Computational time in terms of the parametric queries for LCO\label{fig:LCO_query}}
\end{figure}

\subsection{Three-dimensional fluid flow}

\subsubsection{Description of the analysis}

\begin{figure}[h!]
\centering
\includegraphics[width=0.65\linewidth]{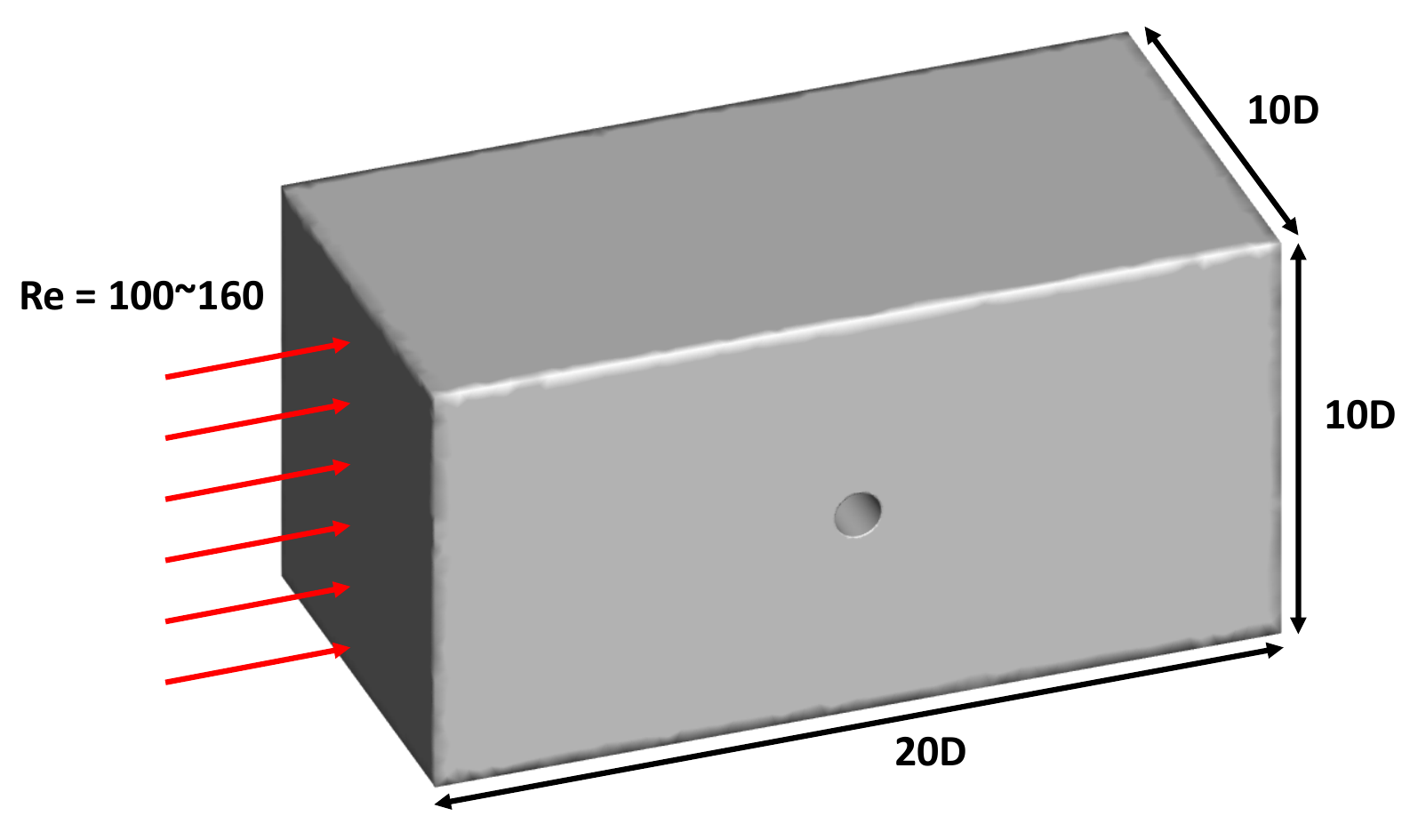}
\caption{Analysis setup of the 3D fluid flow}
\label{fig:fluid_setup}
\end{figure}

Finally, fluid flow surrounding a simple stationary three-dimensional (3D) cylinder is analyzed. The analysis of the 3D fluid serves to demonstrate the use of the proposed framework to analyze a system with a significant number of DOFs. A 3D cylinder with a diameter of 1 m was subjected to a uniform inflow, as shown in Fig. \ref{fig:fluid_setup}. Similar to Turek-Hron FSI benchmark, a von K\`{a}rm\`{a}n vortex is formed behind the cylinder. For CFD analysis, a cuboid computational domain of $20m\times10m\times10m$ was discretized into 1,121,000 tetrahedral elements. The Reynolds number of the inflow varied from 100 to 160 in intervals of 10.

The initial FOM samples are obtained by using the ANSYS Navier-Stokes solver and 2s of FOM data are collected in intervals of 0.01 s. Then, the LSH-VAE is trained against the FOM ensemble and interpolation is performed with respect to the parameter.

After LSH-VAE is trained, the latent code representing the targeted parameter is acquired. The target parameter is selected as an unseen inflow Reynolds number of $Re = 125$. The latent code corresponding to $Re = 125$ is acquired by the interpolation shown in Algorithm \ref{alg3}. The interpolated latent code is then decoded and the resultant interpolated flow field is generated.

\subsubsection{Evaluation of the accuracy and efficiency}

\begin{figure*}[h!]
\captionsetup[sub]{font=normal}
\centering

\begin{subfigure}[b]{0.43\textwidth}
%\centering
\includegraphics[width = \textwidth]{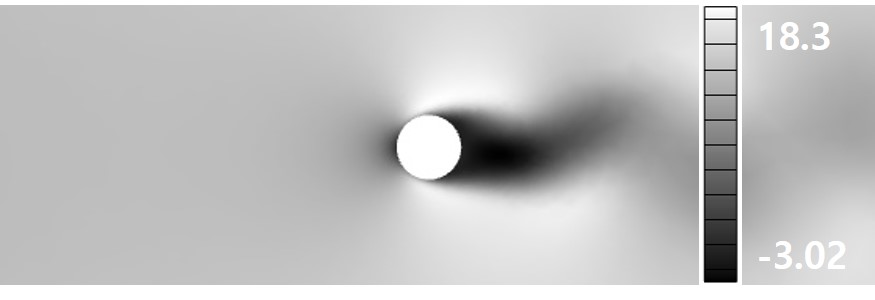}
\caption{LSH-VAE, $u$}
\end{subfigure}
\begin{subfigure}[b]{0.43\textwidth}
\centering
\includegraphics[width = \textwidth]{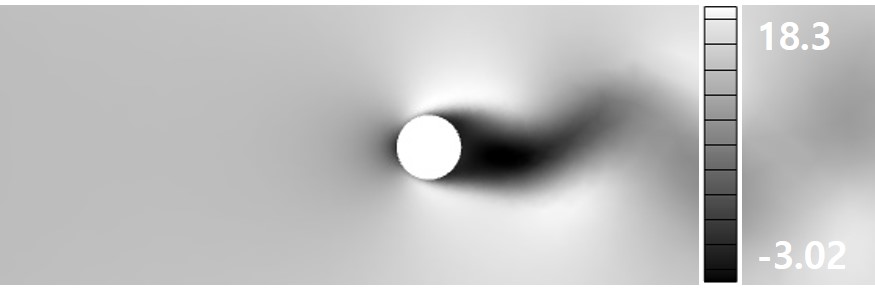}
\caption{FOM, $u$}
\end{subfigure}

\begin{subfigure}[b]{0.43\textwidth}
\centering
\includegraphics[width = \textwidth]{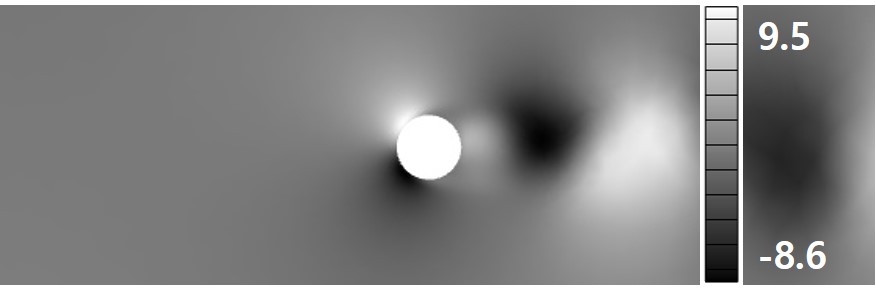}
\caption{LSH-VAE, $v$}
\end{subfigure}
\begin{subfigure}[b]{0.43\textwidth}
\centering
\includegraphics[width = \textwidth]{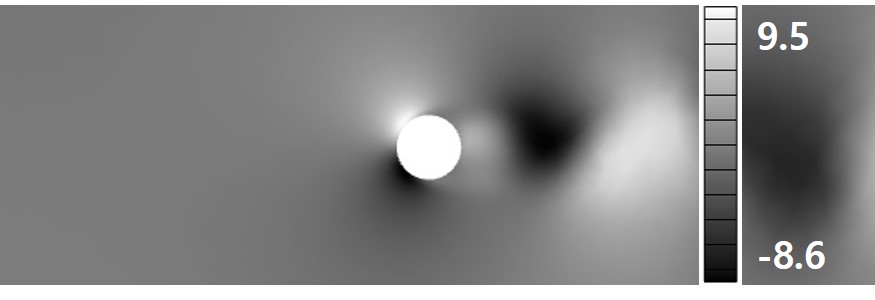}
\caption{FOM, $v$}
\end{subfigure}

\begin{subfigure}[b]{0.43\textwidth}
\centering
\includegraphics[width = \textwidth]{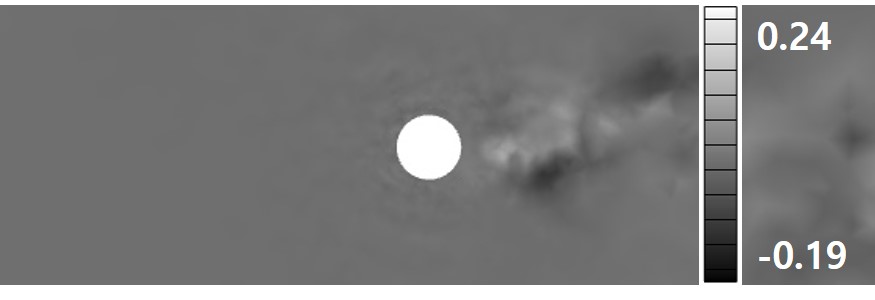}
\caption{LSH-VAE, $w$}
\end{subfigure}
\begin{subfigure}[b]{0.43\textwidth}
\centering
\includegraphics[width = \textwidth]{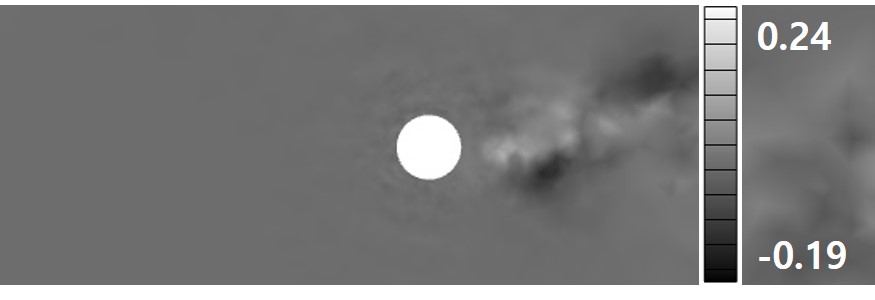}
\caption{FOM, $w$}
\end{subfigure}

\begin{subfigure}[b]{0.43\textwidth}
\centering
\includegraphics[width = \textwidth]{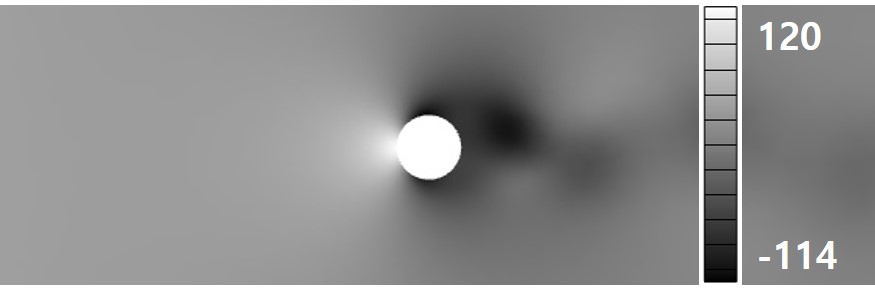}
\caption{LSH-VAE, $p$}
\end{subfigure}
\begin{subfigure}[b]{0.43\textwidth}
\centering
\includegraphics[width = \textwidth]{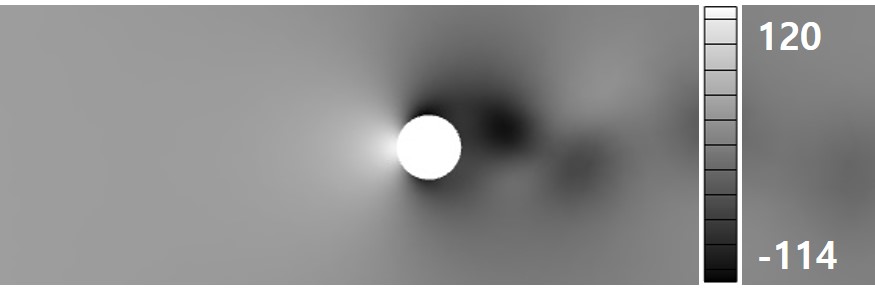}
\caption{FOM, $p$}
\end{subfigure}
\caption{Original and interpolated FSI field for 3D fluid flow at $t=2$ s
}\label{fig_3D_field}
\end{figure*}

The accuracy of LSH-VAE is assessed by comparing the results of ROM with those obtained using FOM. In this case, four physical variables, $u, v, w$, and $p$ are considered for the interpolation. Using the interpolated variables, the interpolated flow field is generated. The interpolated and original flow fields are displayed in Fig. \ref{fig_3D_field}. 

\begin{figure*}[h!]
\captionsetup[sub]{font=normal}
\centering

\begin{subfigure}[b]{0.7\textwidth}
%\centering
\includegraphics[width = \textwidth]{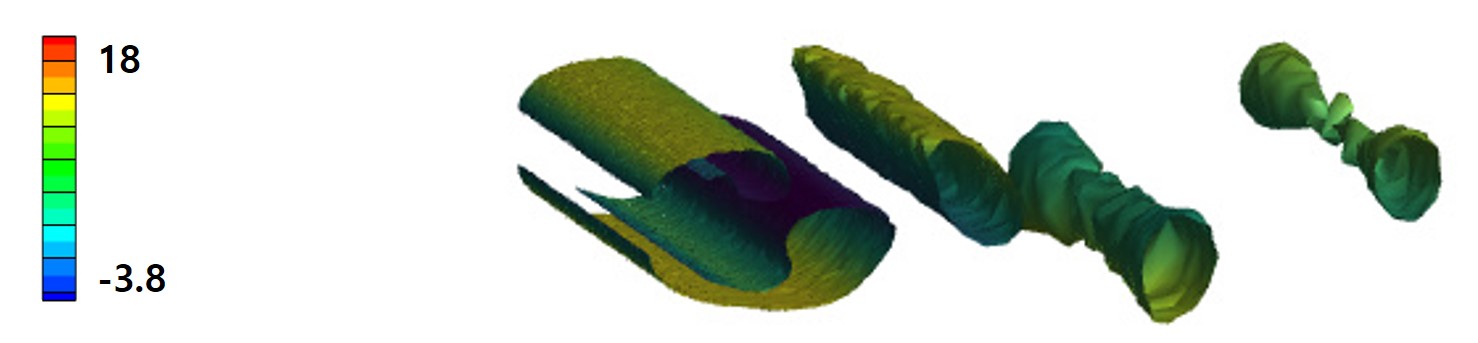}
\caption{LSH-VAE, $Q=0.02,$$u$}
\end{subfigure}
\begin{subfigure}[b]{0.7\textwidth}
\centering
\includegraphics[width = \textwidth]{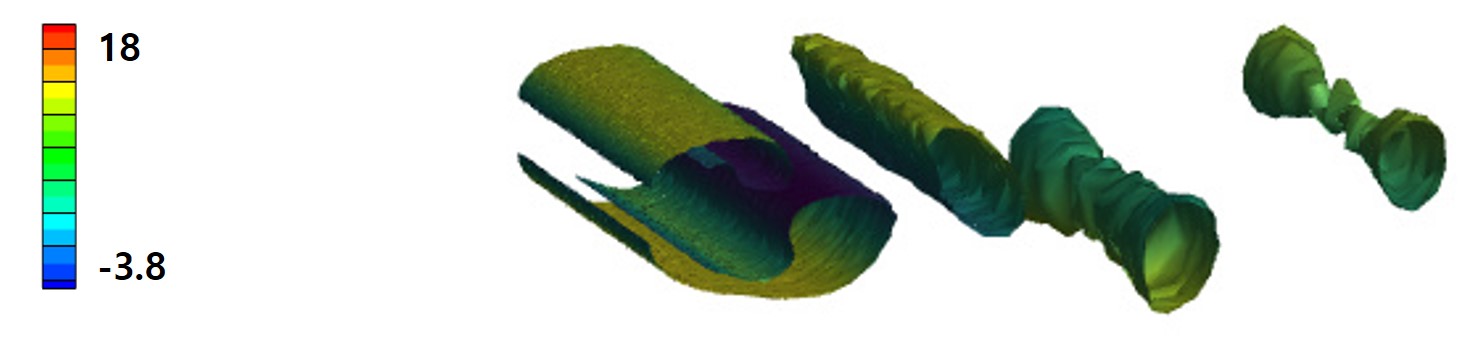}
\caption{FOM, $Q=0.02,$$u$}
\end{subfigure}

\begin{subfigure}[b]{0.7\textwidth}
\centering
\includegraphics[width = \textwidth]{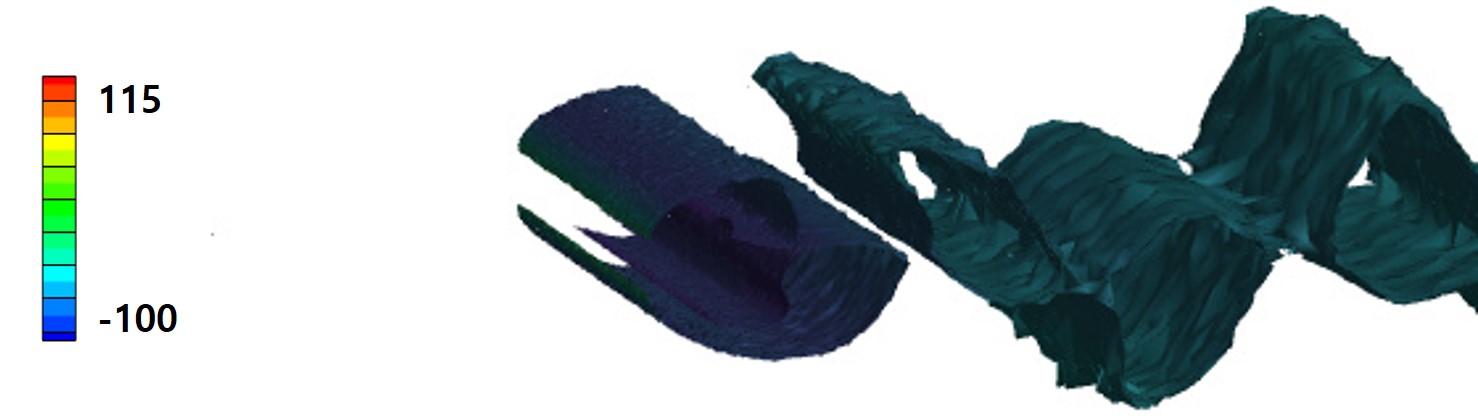}
\caption{LSH-VAE, $Q=0,$$p$}
\end{subfigure}
\begin{subfigure}[b]{0.7\textwidth}
\centering
\includegraphics[width = \textwidth]{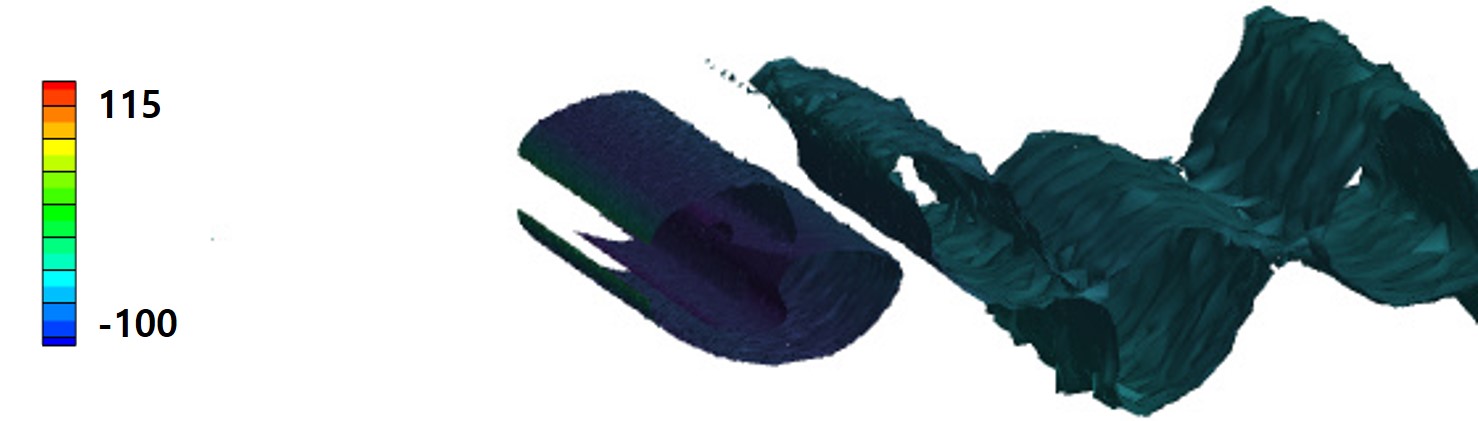}
\caption{FOM, $Q=0,$$p$}
\end{subfigure}

\caption{Original and interpolated flow fields: iso-surface of 3D fluid flow at $t=1.5s$}\label{fig_3D_Q}
\end{figure*}

The interpolated flow field constructed by LSH-VAE is found to be quite accurate. Particularly, the velocity in z-direction, $w$, is accurately interpolated even though $w$ exhibits quite a complex response. As the initial physical variables are interpolated well, the relationship between the variables is inspected. Comparison against CAE and $\beta$-VAE is not conducted in this case as the large number of DOF caused instability of the networks. Instead, the normalized Q-criterion is considered to assess whether the interpolated flow field preserves its vorticity. In Fig.\ref{fig_3D_Q}, the normalized Q-criterion is obtained using the interpolated variables shown in Fig. \ref{fig_3D_field}. Figure \ref{fig_3D_Q} shows the iso-surface generated based on the normalized Q-criterion. The iso-surface is colored by u-velocity and pressure for visualization.

The good agreement in terms of the Q-criterion indicates that LSH-VAE interpolates the direct variables sufficiently well such that the relationship between variables may be well preserved. 

Lastly, the efficiency of the present framework is assessed. The computational time required for each stage is listed in Table \ref{tab:3D}. The offline FOM computation requires 193.7 h including the seven initial FOM samples. LSH-VAE training requires 11.3 h resulting in a total offline stage of 205.0 h. For the online stage, variable reconstruction and writing to disk requires the most time as it required 2.02 h. The proposed framework exhibits a speed-up factor of 14 for each unseen parametric estimation. 
The expected computational time in terms of queries is as shown in Fig. \ref{fig:LCO_query}.

\begin{center}
\begin{table}[h!]
\centering
\caption{Computational time requirements for 3D fluid flow\label{tab:3D}}
\begin{tabular}{llc}
\toprule
&Procedure &Computational time[h]\\ 
\midrule
\multirow{3}{*}{Offline} & FOM & 193.7\\ 
& {Algorithm 1} & 11.3\\
\cmidrule{2-3} 
& Total offline stage & 205.0\\ 
\midrule
\multirow{3}{*}{Online} & Algorithm 2& $<$0.01\\ 
& Variable reconstruction/write & 2.02\\ 
\cmidrule{2-3} 
& Total online stage & 2.02\\ \midrule
&Total sum & 207.0\\ 
%\multicolumn{2}{l}{Total sum} & 82.7\\ 
\bottomrule
\end{tabular}
\end{table}
\end{center}

\begin{figure}[h!]
\centerline{\includegraphics[scale=0.5]{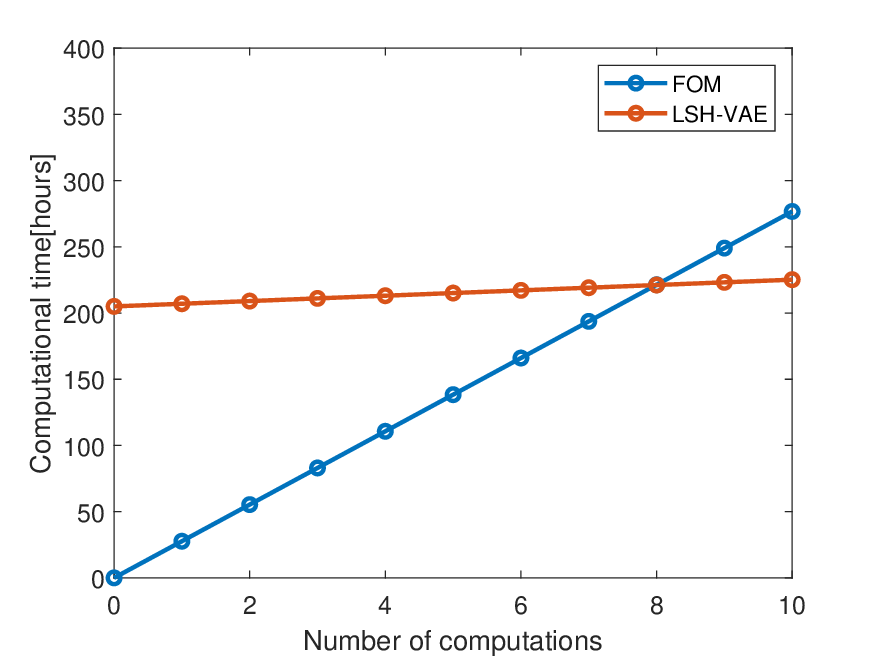}}
\caption{Computational time in terms of the parametric queries for the 3D fluid flow\label{fig:LCO_query}}
\end{figure}

\section{Conclusions}\label{section:conclusion}

This paper proposes a nonlinear data-driven parametric MOR framework based on a neural network. The present framework adopts a novel neural network, LSH-VAE, to perform parametric MOR and interpolation. The present validations demonstrates that the LSH-VAE is capable of the parametric interpolation of dynamic system while significantly reducing the computational time. The following results are obtained in this study.

\begin{itemize}
\item{A novel machine-learning method, LSH-VAE, is developed for nonlinear MOR and the parametric interpolation of nonlinear, dynamic systems.}
\item{LSH-VAE is assessed on three nonlinear and multiphysics dynamic systems with many DOFs. The proposed framework is proven to be accurate and to significantly reduce the computational time.}
\item{Compared against the existing nonlinear MOR methods, convolutional autoencoder and $\beta$-VAE, LSH-VAE demonstrates significantly higher accuracy.}
\end{itemize}

The performance of LSH-VAE is assessed on three nonlinear dynamic systems: FSI benchmark, LCO, and three-dimensional flow. For all of the systems, LSH-VAE is capable of constructing an accurate parametric ROM. Especially, LSH-VAE exhibited a significantly enhanced accuracy compared to CAE and $\beta$-VAE. Also, LSH-VAE is found to be effective as not only did it interpolate the variables well, but it also interpolated the vorticity with high accuracy, which is embedded in the patterns of variables. Upon the accurate parametric MOR, LSH-VAE exhibites a speed-up factor of 990, 660, and 14 respectively. 

Such results are possible owing to the improvements in the LSH-VAE. First, it adopts a hierarchical structure that enables a much deeper and more stable network. Second, it adopts a hybrid weighted loss function consisting of mean-squared error and KL divergence. The use of mean-squared error improved the performance against continuous datasets while the hybrid weights reduced posterior collapse. Lastly, the use of slerp interpolation instead of linear interpolation in the latent space significantly enhanced the interpolation quality following the complex latent manifolds.

However, there still exist a few challenges to be dealt with. First, LSH-VAE may require a significant amount of video random access memory (VRAM) if it is incorporated with an extensive number of DOF. The excessive VRAM requirement stems from its deep structure. By adopting a deep structure, LSH-VAE is capable of generating an expressive result at the cost of training an extensive number of learnable nodes. The excessive VRAM requirements necessitate limiting the batch size for the 3D fluid flow example. Yet, VRAM limitations may be alleviated by adopting parallel computing and utilizing many GPUs. Splitting the DOFs into several groups and merging them after interpolation may also be considered as a solution. Second, extrapolation is limited in the proposed framework. Accurate extrapolation would require dense sampling in the parametric space. However, the construction of ROM with sufficiently dense sampling accompanied by an effective latent manifold tracking method would make reasonable extrapolation viable. Finally, the effectiveness of the proposed framework decreases as the FOM becomes simpler and increasing DOFs are involved. An example of this tendency is demonstrated in the 3D fluid flow example where the speed-up factor diminished to 14 compared to 990 and 660 in the previous cases.

In the future, the plan is to extend the evaluation of the proposed framework to various multiphysics problems such as the analysis of the heat-structure systems. Considering that the present framework is purely data-driven, LSH-VAE is expected to be used in its current form. In addition, multi-parametric analysis coupled with sampling algorithms such as Latin hypercube will be attempted by adopting conditional tokens in the latent space.

\backmatter

\bmhead{Acknowledgments}

This research was supported by the Basic Science Research Program through the National Research Foundation of Korea (NRF), funded by the Ministry of Science, ICT and Future Planning (2023R1A2C1007352).

\section*{Declarations}

The authors declare that they have no conflict of interest.


\bibliography{Refs}% common bib file


\end{document}